\documentclass[10pt]{article}

\usepackage[preprint]{tmlr}

\usepackage{iftex}
\ifPDFTeX
  \usepackage[T1]{fontenc}
  \usepackage[utf8]{inputenc}
\fi
\usepackage{lmodern}
\usepackage{microtype}
\usepackage{amsmath,amssymb,amsfonts}
\usepackage{graphicx}
\usepackage{textcomp}
\usepackage{xcolor}
\usepackage{booktabs}
\usepackage{array}
\usepackage{needspace}
\usepackage{multirow}
\usepackage{subcaption}
\usepackage[section]{placeins}
\usepackage{float}
\usepackage{flafter}

\definecolor{tmlrlink}{rgb}{0.0,0.4,0.75}
\usepackage[colorlinks=true,
            linkcolor=tmlrlink,
            citecolor=tmlrlink,
            urlcolor=tmlrlink,
            bookmarks=true]{hyperref}
\hypersetup{pdftitle={Anchor-and-Harvest Curriculum: A Fully Unmasked Non-LLM Curriculum for Craftax},pdfauthor={Jiayu Zhu, Luyuan Li, Yuyuan Shi, Hanyi Dong}}
\usepackage{bookmark}

\newcommand{\secref}[1]{\S\ref{#1}}
\newcommand{\apxref}[1]{Appendix~\ref{#1}}

\newcolumntype{P}[1]{>{\raggedright\arraybackslash}p{#1}}


\title{Anchor-and-Harvest Curriculum:\\
  A Fully Unmasked Non-LLM Curriculum for Craftax}
\author{%
  {\setlength{\tabcolsep}{3pt}%
  \begin{tabular}[t]{@{}*{4}{>{\raggedright\arraybackslash}p{\dimexpr0.25\textwidth-1.5\tabcolsep\relax}}@{}}
  \name Jiayu Zhu\thanks{All authors contributed equally as first authors.} &
  \name Luyuan Li\footnotemark[1] &
  \name Yuyuan Shi\footnotemark[1] &
  \name Hanyi Dong\footnotemark[1] \\
  \addr Brown University\newline jiayu\_zhu@brown.edu &
  \addr York University\newline mas0n@my.yorku.ca &
  \addr Sun Yat-sen University\newline shiyy57@mail2.sysu.edu.cn &
  \addr The Chinese University of Hong Kong, Shenzhen\newline 123020084@link.cuhk.edu.cn
  \end{tabular}}}
\date{}
\begin{document}
\raggedbottom
\maketitle

\begin{abstract}
We introduce Anchor-and-Harvest Curriculum (AHC), a domain-informed, rule-based training system for Craftax-Full. AHC combines achievement-oriented tasks, harvested initial-state progress, difficulty adaptation, bounded selection, and original-task practice. It uses no LLM inference and keeps all 43 actions available. At 2,005,401,600 PPO rollout transitions, three primary AHC runs obtain native-task return $46.70\pm0.57$ (sample SD), compared with $41.91\pm1.01$ for fresh native PPO-GTrXL with the same student architecture and evaluation protocol. Evaluation uses ten stochastic rollouts on each of 1024 fixed worlds per model. An additional cohort of three AHC runs reaches $46.08\pm1.23$. Two single-component ablations retain similar mean returns, while jointly freezing difficulty and removing inherited progress gives $45.14\pm2.79$. All-67-achievement profiles show that most of the return difference comes from Intermediate achievements, while the deepest-floor achievements remain unsolved. A Skill-Preflight diagnosis with a 14B teacher shows how prerequisite and floor-access constraints restrict practice. AHC and three 14B diagnostic configurations reach aggregate returns near the reported 80B DiCode mean of 45.88 and below the 235B mean of 48.55. These results show that an explicit curriculum without LLM inference can reach returns comparable to those reported for LLM-generated curricula on Craftax-Full, without action masking.
\end{abstract}

\section{Introduction}
\label{sec:intro}
Curriculum learning in open-ended environments must decide what to practise,
where practice begins, and when its difficulty should change. Craftax-Full
\citep{matthews2024craftax} makes these decisions consequential: reaching
a new floor and learning the abilities needed there are distinct challenges.
LLM-written curricula offer one approach. Dreaming-in-Code (DiCode)
\citep{mitsides2026dicode} generates executable levels and reports strong
performance with a large teacher. We investigate how far an explicit,
domain-informed controller can go without LLM generation or action filtering.

AHC enumerates achievement-oriented tasks, harvests progress signatures from
session-ending training states, and combines these signatures with
procedurally generated worlds to construct starting states for curriculum
practice. A controller adapts task difficulty and allocates training
environments across curriculum tasks and an original-task anchor within a
fixed PPO rollout budget; the anchor uses ordinary environment resets.
Evaluation uses the native Craftax dynamics and rewards.
The controller draws on engine knowledge of recipes, floors, and achievement
dependencies, which takes the place of the domain knowledge an LLM teacher
would otherwise supply.

Our empirical question is whether this combined task-and-state training
procedure yields better original-task outcomes than fresh native PPO-GTrXL
at the same PPO rollout budget. We therefore compare complete training
systems, and use ablations to probe two of AHC's mechanisms.

The method is motivated by a focused analysis of Skill-Preflight, a constraint
stack added to a 14B DiCode teacher. Under observed mastery profiles,
prerequisite rules leave 27 downstream targets ineligible for direct
scheduling. A separate floor clause limits deep-floor practice. Removing the
floor clause in the Surgical configuration changes the distribution of
deep-floor practice, and all three runs exceed the four-skill thresholds
defined in Section~\ref{sec:system}. These findings motivate explicit
control of practice access. The diagnosis also shows why we report
achievement-level outcomes alongside aggregate return: the aggregate-return
comparison between three base15 and three longStack runs is inconclusive,
yet the lowest base15 success rate exceeds the highest longStack rate on all
four reported deep-skill outcomes over the last ten checkpoints, and on
three of the four at checkpoint 15300.

Our contributions are:
\begin{enumerate}\itemsep2pt
\item \textbf{An explicit non-LLM curriculum system.} We specify AHC's
progress reuse, difficulty adaptation, eligibility, selection, retirement,
and original-task anchoring, with an implementation-level description of
their interaction. AHC builds on established learnability scores and
rank-based replay; the contribution is the integrated procedure.
\item \textbf{A budget-indexed fully unmasked Craftax result.} Three primary AHC runs achieve $46.70\pm0.57$ at 2.0054016B PPO rollout transitions and $38.69\pm2.26$ at 0.9961472B. Fresh native PPO-GTrXL, using the same student and evaluation protocol, reaches $41.91\pm1.01$ and $36.63\pm1.42$ at these budgets. A further three-run AHC cohort, common-endpoint single-rollout comparisons, and three ablation arms with three training seeds each (two individual interventions and their joint intervention) complete the evaluation.
\item \textbf{A Skill-Preflight diagnosis of constrained practice.} With a
14B teacher, source rules and interventions show how task eligibility and
floor access restrict practice, and why aggregate return alone does not
describe capability allocation. The repairF, shufGraph, and Surgical
configurations reach aggregate returns numerically near the published DiCode
mean of 45.88 for an 80B teacher \citep{mitsides2026dicode}, while their
capability profiles differ
(Appendix~\ref{apx:return-context}). The diagnosis supplies the design
hypotheses behind AHC.
\end{enumerate}

All AHC, ablation, and native-PPO checkpoints, together with one additional fixed-difficulty checkpoint, are evaluated with the same repeated-world protocol: 37 checkpoints and 378,880 episodes in total (Appendix~\ref{apx:binding-status}).

\begin{figure}[htb]
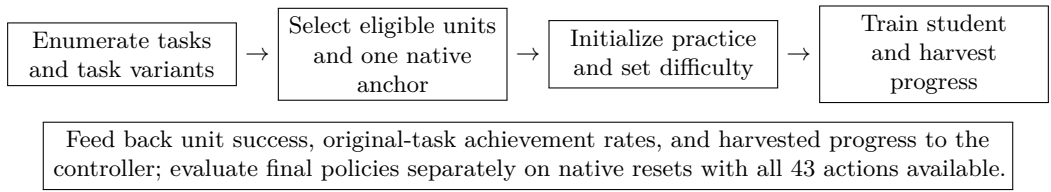

\centering\small
\setlength{\tabcolsep}{2pt}
\begin{tabular}{c@{\ $\rightarrow$\ }c@{\ $\rightarrow$\ }c@{\ $\rightarrow$\ }c}
\fbox{\parbox{2.8cm}{\centering Enumerate tasks\\and task variants}} &
\fbox{\parbox{2.8cm}{\centering Select eligible units\\and one native anchor}} &
\fbox{\parbox{2.8cm}{\centering Initialize practice\\and set difficulty}} &
\fbox{\parbox{2.8cm}{\centering Train student\\and harvest progress}}
\end{tabular}
\par\medskip
\fbox{\parbox{12.6cm}{\centering Feed back unit success, original-task achievement
rates, and harvested progress to the controller; evaluate final policies
separately on native resets with all 43 actions available.}}
\caption{AHC's data flow. The motivating diagnosis concerns restricted task
access, missing deep practice, and limited information in aggregate return.}
\label{fig:overview}
\end{figure}

\section{Setting and Evaluation Protocol}
\label{sec:system}
\paragraph{Environment and student.}
Primary experiments use Craftax-Symbolic-v1 in Craftax 1.4.5, the symbolic
Craftax-Full environment \citep{matthews2024craftax,hafner2022crafter}.
AHC retains the PPO-GTrXL student of the DiCode implementation
\citep{schulman2017ppo,parisotto2020gtrxl,mitsides2026dicode} and replaces
its LLM curriculum generation and control. AHC performs 200 initialization
updates followed by 151 sessions of 100 PPO updates each. With 1,024
parallel environments and 128 rollout steps per update, this yields exactly
2,005,401,600 PPO rollout transitions, excluding evaluation.
Signature harvesting and controller success statistics reuse training
rollouts without a separate environment-stepping pass. Self-imitation
learning is off.

\paragraph{Native and fully unmasked evaluation.}
Native evaluation uses original dynamics, procedural resets, and native
rewards, without harvested initial states or synthetic-task reward changes.
All 43 actions remain available during AHC training and evaluation, with
no ground-truth or predicted feasibility filter. The policy is conditioned
on a 67-dimensional multi-hot achievement indicator; for the native full
task all 67 entries are one.

\paragraph{Endpoints and uncertainty.}
The AHC endpoint is checkpoint 15300 for training seeds 0--2. The primary
estimate uses ten stochastic rollouts on each of the 1024 worlds generated
with evaluation seed 0: 10,240 episodes per model. For every repeat, the
evaluator reconstructs the same reset batch; a repeat-specific play key then
supplies the random streams for action sampling and environment transitions.
Evaluation uses ordinary vectorized batching,
an 8192-step cap, and each environment's first episode; cap-length episodes
are included in the denominator. Appendix~\ref{apx:d1} specifies the schedule.
The same evaluator and schedule are applied to AHC checkpoint 7600, the additional
full-AHC executions, the D4 ablations, and native-PPO checkpoints, including
three full-budget runs (Appendix~\ref{apx:unified}).

The earlier single-rollout evaluation on the same 1024 worlds is used
for historical comparisons. Historical Skill-Preflight diagnostics instead
use the within-run average over checkpoints 14400--15300 as their primary
statistic; these tail-window results are reported separately. Because
DiCode reports the value at its final checkpoint \citep{mitsides2026dicode},
each diagnostic run's checkpoint-15300 reading from the same evaluation
summaries is also reported (Appendix~\ref{apx:rates}), and claims that depend
on the reading are stated under both. The direct AHC--base15--longStack
comparison uses the checkpoint-15300 single-rollout value of each run.
Unless marked SE, uncertainty is sample SD across training runs.

\paragraph{Diagnostic comparisons.}
Base15 and longStack share the \texttt{qwen2.5-coder:14b} teacher and
15-parent supply setting. LongStack adds prerequisite scheduling, a mastery
rule, preflight rollouts, the scaffold gate, and its mastered-prerequisite
exemption. Task supply is approximately matched across configurations, with
recorded slot deficits of 1.1\% for baselines and 3.3--5.0\% for
stack-family configurations.
Surgical removes only the floor clause and shares the API-reference ON
marker with longStack; comparisons that also change the API-reference
setting are identified with the graph results.

\paragraph{Skill boundary.}
We say a run \emph{retains} the skill boundary when it strictly exceeds
four success-rate thresholds: 10.26\% for mine entry, 4.54\% for warrior,
0.28\% for archer, and 0.88\% for bat, the tail-window rates of longStack s1. The label summarizes the raw rates,
which remain the primary evidence; the AHC, ablation, and native-PPO counts
are unchanged when all four thresholds are scaled by 0.5--1.5
(Appendix~\ref{apx:all-capabilities}). Historical diagnostic recovery uses
the tail window, and every recovery label is unchanged at checkpoint 15300
(Table~\ref{tab:graph}, Appendix~\ref{apx:rates}). AHC retention is
evaluated at its final endpoint.

\begin{table}[htb]
\centering\small
\caption{Summary of the experimental setting.}
\label{tab:dependencies}
\begin{tabular}{P{3.0cm}P{10.8cm}}
\toprule
Item & Setting \\
\midrule
Environment & Craftax-Symbolic-v1, Craftax 1.4.5 \\
Budget & 2,005,401,600 PPO rollout transitions (15300 updates); harvest and controller statistics reuse these rollouts; evaluation is excluded \\
Endpoints & Update 15300 (final) and 7600 (below 1B); diagnostic tail window 14400--15300 \\
Evaluation & 1024 fixed worlds $\times$ 10 stochastic rollouts, 8192-step cap; historical comparisons use one rollout per world \\
Actions and inputs & 43 actions, no filter; 67-dimensional task indicator \\
Training-state access & Harvested progress, softened dynamics, engine-informed dependency graph \\
\bottomrule
\end{tabular}
\end{table}

\section{Diagnosis: Constraints on Practice and Capability}
\label{sec:diagnosis}
This section motivates AHC's design and shows what an aggregate curriculum
score can conceal.

\subsection{Prerequisite rules and the depth wall}
\label{sec:gate}
\begin{figure}[htb]
\centering
\includegraphics[width=0.94\linewidth]{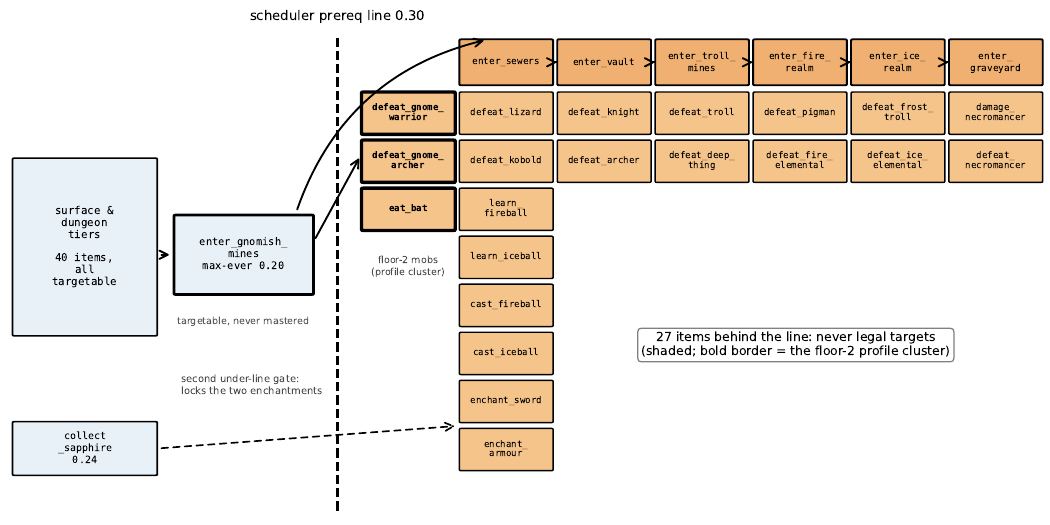}
\caption{The archive lock, drawn from the scheduler's eligibility record (graph fingerprint
\texttt{cfbb1c9a4558}). Eligibility under the true prerequisite graph, over
five first-round arms' end-of-run mastery profiles: an item is a legal target
only when every prerequisite's max-ever mastery clears the scheduler's 0.30
line. \texttt{enter\_gnomish\_mines} is itself targetable but topped out at
0.20 against that line, and the 27 items behind it --- of 67 in the graph, the
three floor-2 cluster mobs among them --- therefore remain ineligible
under these measured profiles. One further under-line gate
(\texttt{collect\_sapphire}) additionally locks the two enchantments inside
the same closure. The mastery profiles come from the first round of
diagnostic runs.}
\label{fig:lock}
\end{figure}

The scheduler admits a target only when each of its prerequisites clears
the max-ever mastery threshold of 0.30. In the five first-round endpoint profiles,
mine entry reaches at most 0.20, leaving 27 downstream targets ineligible
under the original graph (Figure~\ref{fig:lock}). The lock follows from
the graph, the rule, and the measured profiles, not from the difficulty of
the locked skills themselves.

\begin{table}[htb]
\centering\small
\caption{Graph configurations in the diagnostic record. Warrior is the s1
rate (\%) averaged over the last ten checkpoints (tail) and at checkpoint
15300 (final). Recovery counts use three training runs and the
four-skill boundary; they are identical under both readings. RepairF also
changes the API-reference setting relative to longStack.}
\label{tab:graph}
\begin{tabular}{P{2.6cm}P{5.4cm}rrrr}
\toprule
 & & \multicolumn{2}{c}{Warrior (s1)} & \multicolumn{2}{c}{Recovery} \\
\cmidrule(lr){3-4}\cmidrule(lr){5-6}
Configuration & Graph and comparison scope & Tail & Final & Tail & Final \\
\midrule
longStack & Original graph; floor clause on & 4.54 & 6.74 & 0/3 & 0/3 \\
repairF & Repaired graph; API-reference differs & 14.36 & 16.11 & 3/3 & 3/3 \\
shufGraph & Fixed shuffled graph; changes both graph consumers & 14.97 & 16.02 & 3/3 & 3/3 \\
\bottomrule
\end{tabular}
\end{table}

RepairF changes the prerequisite of each of the warrior, archer, and bat
targets from mine entry to dungeon entry, while leaving the other 91 of the
graph's 94 edges unchanged. ShufGraph
uses one fixed degree-preserving rewiring (seed 20260804) for all three
training seeds. Both the scheduler and the scaffold gate consume the
changed graph. The intervention rules, API-reference difference, and gate criterion are
specified in Appendix~\ref{apx:graph-timing}.

\begin{figure}[htb]
\centering
\includegraphics[width=0.94\linewidth]{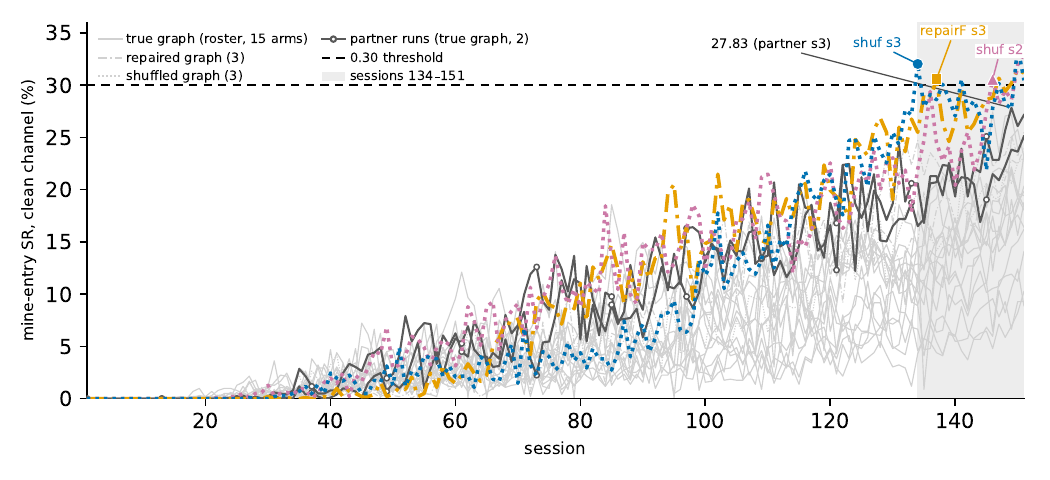}
\caption{Per-session mine-entry success on the held-out clean channel for 21
diagnostic runs and two separately marked partner runs. Line style identifies
the true, repaired, or shuffled prerequisite graph. No true-graph series reaches
the 0.30 scheduler threshold; the closest is partner base2e9 s3 at 27.83\%.
Three altered-graph runs cross late: shufGraph s3 at session 134, repairF s3 at
137, and shufGraph s2 at 146.}
\label{fig:wall}
\medskip
\captionof{table}{Readings from Figure~\ref{fig:wall}'s monitoring trajectories (\%). Peak is the largest session reading across the listed runs; session-151 columns identify individual runs. The 0.30 mastery line corresponds to 30\%. Base2e9 s2/s3 are the separately marked partner runs; breadth08 and cap300 each have one run.}
\label{tab:wall}
\small\setlength{\tabcolsep}{4pt}
\begin{tabular}{llrrrrl}
\toprule
Configuration & Graph & Peak & \multicolumn{3}{c}{Session 151} & First crossing \\
\cmidrule(lr){4-6}
 & & & s1 & s2 & s3 & Run / session \\
\midrule
base15 & Original & 27.44 & 24.90 & 16.41 & 19.82 & None \\
base2e9 & Original & 27.83 & 20.31 & 25.10 & 27.15 & None \\
longStack & Original & 16.21 & 14.06 & 9.28 & 16.21 & None \\
gateOff & Original & 19.24 & 18.07 & 18.75 & 10.25 & None \\
Surgical & Original & 25.78 & 23.14 & 18.36 & 23.34 & None \\
repairF & Repaired & 31.54 & 23.93 & 25.49 & 31.15 & s3 / 137 \\
shufGraph & Shuffled & 33.79 & 23.05 & 33.01 & 30.47 & s3 / 134; s2 / 146 \\
breadth08 & Original & 12.40 & 8.89 & -- & -- & None \\
cap300 & Original & 22.17 & 19.43 & -- & -- & None \\
\bottomrule
\end{tabular}
\end{figure}

Across the 21 registered diagnostic runs and two separately marked,
environment-aligned partner runs, no original-graph series reaches the
0.30 mine-entry line in 151 training-period clean-channel readings. The
closest is 27.83\%. Only shufGraph s2/s3 and repairF s3 cross, all after
session 133 (Figure~\ref{fig:wall}; Table~\ref{tab:wall} gives the readings
by configuration). Altered-graph configurations recover the skill boundary
in all runs under either reading.

\subsection{Floor access is consequential}
\label{sec:floor}
The scaffold gate rejects shortcuts through unmet prerequisites, including
starting on a floor that skips the required entry chain. Surgical removes
only this floor clause while retaining the original prerequisite graph.
In the s1 contrast, warrior tasks starting on floor 2 or deeper change from
0 of 25 to 32 of 33; floor-0 mass falls by 72.13 percentage points and
floor-2 mass rises by 34.84 points. The source change therefore alters
direct practice access, and the skill boundary recovers in all three
Surgical runs under either reading.

The paired surgical-minus-longStack return changes are positive in every
pair: $+0.131$, $+2.101$, and $+2.146$ over the last ten checkpoints (mean
$+1.460$, sample SD $1.151$), and $+1.900$, $+2.502$, and $+1.673$ at
checkpoint 15300 (mean $+2.025$, sample SD $0.429$). Original-graph
scheduling can still exclude direct downstream targets after the floor
clause is removed; warrior goals may arise within mines-directed levels.
Task eligibility, practice access, and acquired capability must therefore
be distinguished. The graph-exemption timing figure is in
\apxref{apx:graph-timing}.

\subsection{Selection can compound an access bottleneck}
The inherited learnability term $p(1-p)$ is small near both failure and
saturation, and only A/B-graded tasks can become parents. First-training
passers become breedable at 58\%, versus 7.5\% for first failures in the
archived task record. Tasks that fail early therefore rarely become parents,
which motivates fresh-task admission, difficulty adjustment, and stale-task
reconsideration in AHC.

\subsection{Aggregate return hides capability differences}
\label{sec:profile}
\begin{table}[htb]
\centering\small
\caption{Diagnostic LLM curriculum comparison under two readings of the same
evaluation summaries: the mean of the last ten checkpoints (primary) and the
final checkpoint 15300, which is DiCode's reporting convention. Uncertainty
is sample SD across three runs. The checkpoint-15300 values are those
compared with AHC in Table~\ref{tab:local-endpoint}; the last-ten means are
not used as comparators for AHC.}
\label{tab:returns}
\begin{tabular}{lcccc}
\toprule
 & \multicolumn{2}{c}{Last ten checkpoints} & \multicolumn{2}{c}{Checkpoint 15300} \\
\cmidrule(lr){2-3}\cmidrule(lr){4-5}
Configuration & Return & Individual runs & Return & Individual runs \\
\midrule
base15 & $43.929\pm1.309$ & 45.016 / 42.476 / 44.296 & $44.509\pm1.780$ & 46.024 / 42.549 / 44.955 \\
longStack & $44.014\pm0.795$ & 44.726 / 43.157 / 44.159 & $44.589\pm0.909$ & 45.016 / 43.545 / 45.206 \\
\bottomrule
\end{tabular}
\end{table}
The base15-minus-longStack difference is $-0.084362$, with standard error
$0.883950$, Welch degrees of freedom 3.2982, and 95\% interval
$[-2.75893,2.59020]$; at checkpoint 15300 it is $-0.079$, with 95\% interval
$[-3.77,3.61]$. The aggregate comparison is inconclusive under either
reading. By contrast, over the last ten checkpoints every base15 run has a
higher success rate than every longStack run on each of the four reported
deep-skill outcomes; at checkpoint 15300 this complete separation persists
for warrior, archer, and bat, while one cross-arm pair does not follow this
ordering for mine entry (Table~\ref{tab:profile}).

\begin{table}[htb]
\centering\small
\caption{Capability profiles (rates in \%; three runs per arm)
over the last ten checkpoints (primary) and at checkpoint 15300. In the tail
window, the smallest base15 value exceeds the largest longStack value for
every listed skill. At checkpoint 15300 this holds for warrior, archer, and
bat but not for mine entry, where base15 s2 reaches 13.96 against 14.94 for
longStack s3. Per-run rates are in Tables~\ref{tab:perseed}
and~\ref{tab:perseed-final}.}
\label{tab:profile}
\begin{tabular}{lcccc}
\toprule
 & \multicolumn{2}{c}{Last ten checkpoints} & \multicolumn{2}{c}{Checkpoint 15300} \\
\cmidrule(lr){2-3}\cmidrule(lr){4-5}
Skill & base15 & longStack & base15 & longStack \\
\midrule
Mine entry & $19.10\pm3.95$ & $9.64\pm1.44$ & $19.99\pm5.52$ & $11.78\pm3.51$ \\
Defeat gnome warrior & $15.12\pm3.64$ & $1.55\pm2.59$ & $16.67\pm4.83$ & $2.25\pm3.89$ \\
Defeat gnome archer & $8.96\pm3.15$ & $0.094\pm0.164$ & $9.96\pm3.98$ & $0.065\pm0.113$ \\
Eat bat & $14.08\pm3.63$ & $0.50\pm0.45$ & $15.07\pm5.01$ & $0.81\pm0.75$ \\
\bottomrule
\end{tabular}
\end{table}

For scale, under an unpaired rank-label permutation with six exchangeable
runs, complete three-versus-three separation in a given direction occupies
one of $\binom{6}{3}=20$ labelings (one-sided 0.05; two-sided 0.10). This
holds for four skills in the tail window and three at checkpoint 15300.

\section{The Anchor-and-Harvest Curriculum}
\label{sec:mechanical}
AHC implements explicit control over task supply, practice initialization,
difficulty, and allocation. Table~\ref{tab:design} connects these operations
to the diagnostic observations.

\paragraph{Candidates and reusable progress.}
AHC disables the foundation-model generation worker and enumerates
achievement, family, and depth units from Craftax's fixed achievement set.
Table~\ref{tab:unit-definitions} distinguishes task labels from physical floors
and the continuous difficulty parameter. A \emph{family} is one of twelve
crafting recipes; a \emph{depth} unit targets entry to a specified floor;
an ordinary achievement unit covers one of the remaining 47 achievements.
The active base task universe comprises 147 units before variants are
added: 24 family L1/L2 units, 64 depth L1/L2 units, 12 family L4 units, and
47 ordinary achievement units; a candidate must
still pass the supply and eligibility checks below. Every two sessions, AHC
proposes up to three task variants using prefill, softening, or ore-placement
operators. A bank stores progress signatures collected from visited
states. The assembler combines these signatures with a bounded pool of
official, procedurally generated worlds indexed by floor. This reuses
achieved progress to initialize practice, rather than replaying gradients.
Each curriculum task targets one achievement. An \texttt{OriginalTask}
anchor always supplies ordinary resets; empty curriculum supply leaves an
anchor-only session.

\begin{table}[htb]
\centering\small
\caption{Diagnostic observations and AHC design responses. AHC retains
hard eligibility rules alongside its soft scores.}
\label{tab:design}
\begin{tabular}{P{5.3cm}P{8.6cm}}
\toprule
Observation & Design response \\
\midrule
Some downstream targets are not directly schedulable & Use the graph for downstream value and explicit, separately specified eligibility/frontier rules. \\
Direct deep-floor practice can be restricted & Represent harvested floor progress through procedurally generated, floor-indexed worlds. \\
Difficult tasks can lose selection or parent opportunities & Admit fresh eligible units, mix rank priority with staleness, and adapt task difficulty. \\
Scalar return conceals skill differences & Monitor the original task and report raw held-out skill rates alongside return. \\
\bottomrule
\end{tabular}
\end{table}

\begin{table}[htb]
\centering\small
\caption{Active base task definitions. $L$ is an implementation label, $f$ a
physical floor index (surface $f=0$), and $\sigma$ a dynamics parameter.
Bank availability, prerequisite checks, retirement, and selection remain
additional conditions; Appendix~\ref{apx:ahc} gives the exceptions.}
\label{tab:unit-definitions}
\begin{tabular}{P{2.35cm}P{3.45cm}P{4.15cm}P{3.65cm}}
\toprule
Task class & Target & Initialization and dynamics & Candidate qualification \\
\midrule
Family L1/L2 & Craft the recipe's mapped achievement & Harvested signature on its recorded floor; recipe tools/resource support; both base labels have $\sigma=0$ & Usable bank and equipment-chain check; no depth-neighbor gate \\
Depth L1/L2 & Enter floor $f\in\{1,\ldots,8\}$ & Signature from $f-1$, split by gate-progress bin $g\in\{0,3,6,8\}$; L1 native dynamics, L2 starts at $\sigma=1$ & Bank, neighbor unlock and official-feedback frontier cap \\
Ordinary achievement L0 & One of the 47 non-recipe, non-entry achievements & Signature at its harvested floor; no base surgery/toolkit; $\sigma=0$ & Bank and applicable controller checks \\
Family L4 (official-reset level) & Same twelve recipe achievements & Ordinary native reset; no harvested-state initialization or toolkit; $\sigma=0$ & Requires an official-anchor reading; not eligible for fresh-unit priority \\
\bottomrule
\end{tabular}
\end{table}

For example, \texttt{depth:f2g3|L2} targets mine entry ($f=2$), starts
from a floor-1 signature with three to five recorded kills on that floor,
and starts on the annealed dynamics axis. Neither L2 nor $g=3$ denotes
physical floor 2 or 3. Reaching $\sigma=0$ keeps the L2 label and assisted
initialization; the L1 allocation multiplier still does not apply.

We retain a five-part progress signature comprising the floor, per-floor
gate progress, clamped material/tool inventory, armour slots, and equipment.
The signature excludes the visited map, position, timestep, vitals, and
historical achievement flags. We realize it on a generated world at that floor, apply the selected
prefill and task operators, and derive already-earned achievement flags
before training. The target must still be unsolved. Appendix~\ref{apx:state-interface}
gives field bounds, assembly order, and the training reward/termination
interface. This assisted initialization replaces the native reset for
curriculum tasks.

\paragraph{Training reward and termination.}
For an achievement-targeted curriculum task, each step receives the original
Craftax reward. Reaching the target adds a completion bonus of 20 and
ends the episode. Native-task evaluation uses ordinary resets and native
rewards without this bonus; Appendix~\ref{apx:state-interface} gives the
full reward and termination contract.

\paragraph{Eligibility and selection.}
Units require usable bank supply or an official-reset path and pass depth and
prerequisite checks. Depth unlocking uses an easier neighbor's success rate
of at least 0.50, with exceptions for previously activated units and absent
neighbors. Equipment-derived prerequisites require official-anchor success
of at least 0.10. A frontier rule admits at most the next floor beyond the
deepest floor with a smoothed official reading of at least 0.01.
The exact exceptions and active parameter map are in \apxref{apx:ahc}.

For unit $u$ targeting achievement $a$, $n_u$ counts its completed
episodes logged during the final 40 PPO updates of the current session,
and $k_u$ counts those with the target achievement set at termination.
The logging wrapper excludes single-step episodes. Episodes may begin
before the scoring window, but only those completed and logged within it
contribute to either count. The
assembler checks that the target is unset after achievement fixup, so for
these supplied starts the terminal rate estimates the target's $0\to1$
success probability. The controller maintains $\hat p_u\leftarrow0.5(k_u/n_u)+0.5\hat p_u$,
initialized by the first observation and unchanged when $n_u=0$.
For ordinary units the score is
\begin{equation}
s_u=\hat p_u(1-\hat p_u)R(a),\qquad
R(a)=\sum_{b\in\{a\}\cup\mathrm{Desc}(a)}v_b(1-q_b),
\end{equation}
where $q_b$ is the latest original-task achievement rate from the training
anchor in the same 40-update window (zero if missing),
$v_b$ is its native reward weight, and descendants come from the
engine-informed prerequisite graph, which includes hand-specified engine
dependencies.

Never-active eligible units below the official-reset level are admitted first,
in deterministic identifier order, up to 15 units. Remaining seats use
sampling without replacement with weights
\begin{equation}
w_u=0.7\frac{r_u^{-1}}{\sum_v r_v^{-1}}
    +0.3\frac{\max(0,t-\ell_u)}{\sum_v\max(0,t-\ell_v)},
\end{equation}
where $r_u$ is descending score rank and $\ell_u$ the last training session
for that unit. The staleness term is uniform if its denominator is zero.
Rank temperature is fixed at 1.

\paragraph{Difficulty annealing and retirement.}
For softened units, $\sigma\in[0,1]$ interpolates eight environment parameters
between native dynamics ($\sigma=0$) and an easier setting ($\sigma=1$),
including needs depletion, recovery, enemy spawning, health, and damage.
With at least 100 logged completed task episodes in that window,
the mean of the last three
accepted smoothed success readings reaching 0.85 lowers $\sigma$ by 0.08;
a current reading below 0.40 raises it by 0.04, clipped to $[0,1]$.
Successful arrival at $\sigma=0$ or official mastery can retire depth units.
Official statistics also support reactivation of graduated units after
regression. Retirement, hysteresis, and fallback branches are specified in
\apxref{apx:ahc}. At $\sigma=0$, a harvested-state task has native dynamics
but keeps its assisted initial state.

\paragraph{Practice budget and update loop.}
The selected units plus one anchor give at most 16 task classes. Nominal
curriculum/anchor environment shares are 0.8/0.2. L1 units, identified by
their unit labels, get a weight multiplier of 2. Eligible surviving softened
depth units receive a floor-specific reallocation multiplier after other
units retire. The weights are then normalized globally and converted into
integer environment allocations, so final shares differ from the nominal
values. Each session performs 100 PPO
updates with 1024 environments and 128 rollout steps. Its 100-update session
length and 100-episode statistical gate are distinct: the latter applies
per unit within the final 40 updates. Controller feedback comes from training
rollouts, not the frozen held-out evaluation worlds. Afterward, AHC harvests
progress, updates statistics and controller state, and saves that state for
the next session. Initialization adds 200 updates:
$(200+151\times100)\times1024\times128=2{,}005{,}401{,}600$ transitions.
Self-imitation learning is disabled. Task conditioning uses a
67-dimensional achievement indicator; proposal, mutation, embedding,
training, and evaluation make no LLM inference or API request.

\begin{table}[htb]
\centering\small
\caption{AHC session procedure. The total number of transitions is fixed;
selection and reweighting change their allocation.}
\label{tab:ahc-loop}
\begin{tabular}{P{0.8cm}P{13.2cm}}
\toprule
Step & Operation \\
\midrule
1 & Initialize enumerated units, bank, world pool, per-unit statistics, and
session RNG. Perform the 200-update initialization. \\
2 & For each of 151 sessions, create scheduled task variants; retain units
with supply, then apply unlocking, chain, frontier, and retirement rules. \\
3 & Admit fresh eligible units first; fill the remaining 15-unit capacity
by rank/staleness sampling without replacement. Append the original-task anchor. \\
4 & Assemble curriculum states from harvested signatures and official worlds;
apply each unit's difficulty setting. Reweight and normalize environment shares. \\
5 & Run 100 PPO updates. Harvest progress and collect conditional unit success
and original-task achievement readings. \\
6 & Update EMA statistics, difficulty, graduation, parking, and regression
state; persist the bank/controller and proceed to the next session. \\
\bottomrule
\end{tabular}
\end{table}

\section{Experiments}
\label{sec:experiments}
We evaluate AHC at its final and lower-budget checkpoints, compare it with
native PPO trained with the same student, report three ablation arms with
three seeds each, and give historical and externally reported results as
context. AHC, the ablations, and native PPO share the repeated-world
protocol; historical comparisons and the cohort table keep the original
single-rollout protocol.

\paragraph{Shared student and optimization settings.}
The recorded training configuration uses a two-layer, eight-head gated
Transformer with embedding, QKV, and hidden widths of 256, memory length
128, and a 64-step gradient window. PPO uses $\gamma=0.999$, GAE
$\lambda=0.8$, four epochs, eight minibatches, and Adam.
Table~\ref{tab:shared-training} gives the remaining common settings;
Table~\ref{tab:training-differences} separates the active interventions.

\subsection{Internal comparison at two rollout budgets}
\label{sec:budget}
Table~\ref{tab:internal-budget} consolidates the primary AHC, additional AHC, and fresh native PPO-GTrXL cohorts at both budgets.
Each cohort contains three training runs; both checkpoints come from those
same runs. All use the same GTrXL student architecture and native evaluation
protocol, with all 43 actions available.

\begin{table}[htb]
\centering\small
\caption{Core internal comparison. Return is mean $\pm$ sample SD across
three training seeds per cohort. Each checkpoint is evaluated on the same
1024 native worlds with ten stochastic rollouts per world and an 8192-step
cap. Updates 7600 and 15300 correspond to 0.9961472B and 2.0054016B PPO rollout
transitions, respectively.}
\label{tab:internal-budget}
\begin{tabular}{P{5.2cm}ccc}
\toprule
Training cohort & GPU & Update 7600 & Update 15300 \\
\midrule
Primary AHC & L40S & $38.69\pm2.26$ & $46.70\pm0.57$ \\
Additional AHC & A100 & $40.38\pm1.69$ & $46.08\pm1.23$ \\
Fresh native PPO-GTrXL & RTX A6000 & $36.63\pm1.42$ & $41.91\pm1.01$ \\
\bottomrule
\end{tabular}
\end{table}

At the same 0.9961472B rollout budget, mean return is $38.69\pm2.26$ for
primary AHC, $40.38\pm1.69$ for additional AHC, and $36.63\pm1.42$ for fresh native
PPO. The AHC means are 2.06 and 3.76 points above fresh native PPO,
respectively. At 2.0054016B, the means are $46.70\pm0.57$ for primary AHC,
$46.08\pm1.23$ for additional AHC, and $41.91\pm1.01$ for fresh native PPO, gaps
of 4.80 and 4.17 points (Welch 95\% intervals [2.71, 6.88] and [1.58, 6.76];
Table~\ref{tab:return-contrasts}). At update 7600 the primary-cohort
interval includes zero, while at update 15300 both AHC intervals exclude zero.

\begin{figure}[htb]
\centering
\includegraphics[width=\linewidth]{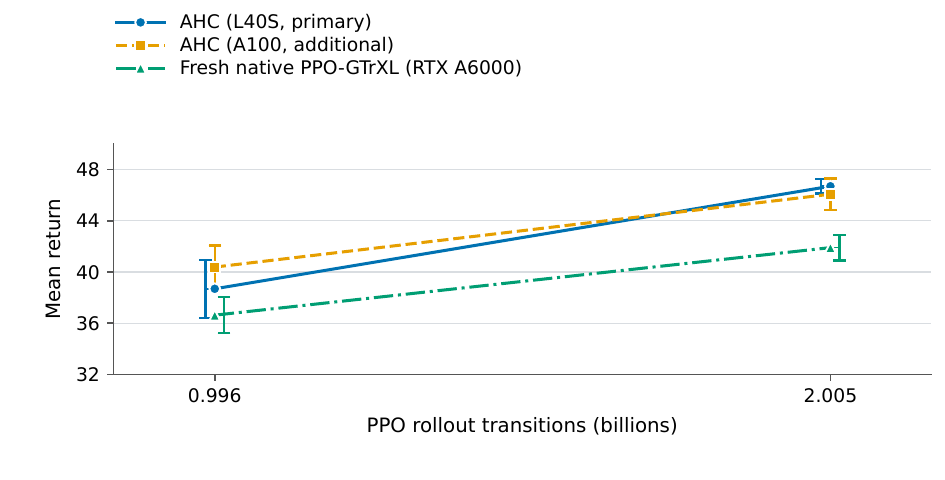}
\caption{Mean return for the three cohorts at the two evaluated PPO rollout budgets. Points are means across three training seeds and error bars are the across-seed sample SD; error-bar stems are offset horizontally for readability. Lines connect the two measured budgets as visual guides. Within each cohort, both points come from the same three training runs.}
\label{fig:budget-comparison}
\end{figure}

Figure~\ref{fig:budget-comparison} displays the same six summaries.
Skill profiles and return composition follow in \secref{sec:pending};
historical and cross-study comparisons are in
Appendix~\ref{apx:return-context}.

\paragraph{Primary endpoint detail.}
Table~\ref{tab:mechanical-endpoint} gives the seed-level repeated-world
results for the primary cohort; its single-rollout readings are used only
for the historical comparison in Appendix~\ref{apx:return-context}.

\begin{table}[htb]
\centering\small
\caption{AHC at the fixed checkpoint-15300 endpoint, using 1024 fixed worlds
and ten stochastic rollouts per world (10,240 episodes/model). Training seeds
are 0--2; the world-generation seed is 0. The final column gives the across-seed mean and sample
SD. Skill rates are percentages; performance is return as a percentage of
the 226-point maximum. Evaluation uses the unfiltered 43-action policy.}

\label{tab:mechanical-endpoint}
\begin{tabular}{lrrrr}
\toprule
metric & seed 0 & seed 1 & seed 2 & mean $\pm$ sample SD \\
\midrule
mean return & 46.945 & 47.105 & 46.051 & $46.700 \pm 0.568$ \\
mean performance (\%) & 20.772 & 20.843 & 20.377 & $20.664 \pm 0.251$ \\
average episode length & 1451.2 & 1583.3 & 1355.3 & $1463.2 \pm 114.5$ \\
\texttt{enter\_gnomish\_mines} (\%) & 28.281 & 27.500 & 23.369 & $26.383 \pm 2.640$ \\
\texttt{defeat\_gnome\_warrior} (\%) & 21.504 & 21.738 & 17.510 & $20.251 \pm 2.377$ \\
\texttt{defeat\_gnome\_archer} (\%) & 13.682 & 14.160 & 13.555 & $13.799 \pm 0.319$ \\
\texttt{eat\_bat} (\%) & 19.014 & 19.756 & 18.193 & $18.988 \pm 0.782$ \\
\bottomrule
\end{tabular}
\end{table}

\subsection{Additional AHC cohort}
\label{sec:cohorts}

The primary full-AHC cohort comprises three runs (seeds 0--2) trained on
L40S GPUs. Three additional AHC runs (seeds 1, 4, and 5), executed later on
A100 GPUs with the same source, pinned dependencies, and resolved
configurations (Appendix~\ref{apx:ahc-repro}), form a second cohort. We
report the two cohorts separately rather than pooling them.
Table~\ref{tab:cohorts} gives their single-rollout readings; repeated-world
results are in Table~\ref{tab:unified}.

\begin{table}[htb]
\centering\small
\caption{Original single-rollout results: one episode on each of 1024 worlds
with an 8192-step cap. The L40S rows form the primary cohort; the A100 rows
are an additional cohort. Boundary marks retention of the
four-skill thresholds in \secref{sec:system}. Seed 4 resumed
training from update 15000 to 15300.}
\label{tab:cohorts}
\begin{tabular}{llccl}
\toprule
Training GPU & Cohort & Seed & Return & Boundary \\
\midrule
L40S & Primary & 0 & 46.861 & Yes \\
L40S & Primary & 1 & 47.268 & Yes \\
L40S & Primary & 2 & 46.134 & Yes \\
\addlinespace
A100 & Additional & 1 & 45.205 & No \\
A100 & Additional & 4 & 47.269 & Yes \\
A100 & Additional & 5 & 44.951 & No \\
\bottomrule
\end{tabular}
\end{table}

Under the single-rollout protocol, all three primary AHC runs and one of
the three additional runs retain the four-skill boundary, and the
repeated-world results preserve these seed-dependent combat differences
(Table~\ref{tab:unified}). In return, the two cohorts differ by $-0.62$ at
update 15300 (Welch 95\% interval [$-3.21$, 1.96];
Table~\ref{tab:return-contrasts}).

\subsection{Matched budgets and run variability}
\label{sec:pending}
Each evaluation record contains the model, world, and repeat identifiers,
return, episode length, terminal flag, and 67 achievement indicators. For
the three primary models, world-clustered within-model return SEs are
0.1894, 0.1897, and 0.1982, well below the across-seed sample SD of 0.5676,
so training-seed variation dominates the uncertainty. Appendix~\ref{apx:d1}
documents the repeated-world design.

The same evaluator is applied to AHC checkpoint 7600
($7600\times1024\times128=996{,}147{,}200$ transitions, the largest saved
budget below 1B), the additional AHC cohort, the nine D4 ablation runs, and
both native-PPO cohorts: the historical guard-stopped runs (at updates 7000
and 8000 and at their stopping points) and the fresh full-budget runs
(Appendix~\ref{apx:unified}). Table~\ref{tab:unified} summarizes the
three-seed groups.

\begin{table}[htb]
\centering\footnotesize
\setlength{\tabcolsep}{3pt}
\caption{Repeated-world evaluation with ten rollouts on each of 1024 fixed
worlds (10,240 episodes per model). Each row is mean $\pm$ sample SD over
three training seeds. I-18 is the unweighted macro success rate across the
18 official Intermediate achievements. Other rates are episode percentages.
W$\mid$entry averages each seed's conditional success rate. Guard-stopped
native-PPO endpoints are historical checkpoints from the original runs;
full-budget native PPO is a separate fresh run set. Training GPUs: L40S for
primary AHC, A100 for A100 AHC, and RTX A6000 for D4 and native-PPO runs.}
\label{tab:unified}
\begin{tabular}{lcccccc}
\toprule
Model & Budget (B) & Return & I-18 & Entry & Warrior & W$\mid$entry \\
\midrule
AHC, checkpoint 7600 (L40S) & 0.996 & $38.69\pm2.26$ & $34.84\pm3.58$ & $7.57\pm1.95$ & $1.51\pm2.48$ & $23.5\pm39.0$ \\
AHC, checkpoint 7600 (A100) & 0.996 & $40.38\pm1.69$ & $37.22\pm2.48$ & $8.05\pm4.22$ & $1.30\pm2.23$ & $19.0\pm32.8$ \\
Fresh native PPO, update 7600 & 0.996 & $36.63\pm1.42$ & $32.62\pm3.06$ & $0.48\pm0.34$ & 0 & 0 \\
\addlinespace
Native PPO, update 7000 (guard-on) & 0.918 & $36.24\pm2.42$ & $31.86\pm4.43$ & $0.47\pm0.36$ & $0.003\pm0.006$ & $0.4\pm0.7$ \\
Native PPO, update 8000 (guard-on) & 1.049 & $38.23\pm1.18$ & $35.49\pm1.88$ & $0.74\pm0.35$ & 0 & 0 \\
\addlinespace
L40S AHC, checkpoint 15300 & 2.005 & $46.70\pm0.57$ & $48.18\pm0.87$ & $26.38\pm2.64$ & $20.25\pm2.38$ & $76.7\pm2.1$ \\
AHC, A100 executions & 2.005 & $46.08\pm1.23$ & $46.77\pm2.06$ & $23.88\pm5.15$ & $8.80\pm13.70$ & $30.4\pm45.3$ \\
D4-A: fixed difficulty & 2.005 & $46.10\pm1.18$ & $47.11\pm1.91$ & $22.71\pm4.45$ & $16.05\pm6.18$ & $69.1\pm15.1$ \\
D4-H: no progress init. & 2.005 & $46.42\pm1.55$ & $48.81\pm1.57$ & $30.79\pm3.02$ & $15.99\pm14.00$ & $53.0\pm45.9$ \\
D4-AH: joint intervention & 2.005 & $45.14\pm2.79$ & $46.13\pm4.78$ & $18.73\pm14.80$ & $9.66\pm16.71$ & $28.0\pm48.2$ \\
Native PPO, guard-stopped & 1.78--1.97 & $41.40\pm1.00$ & $40.79\pm1.75$ & $4.73\pm2.50$ & 0 & 0 \\
Fresh native PPO, full budget & 2.005 & $41.91\pm1.01$ & $41.49\pm1.61$ & $5.43\pm2.81$ & 0 & 0 \\
\bottomrule
\end{tabular}
\end{table}

\paragraph{Near 1B.}
At the shared 7600-update budget (0.9961472B rollout transitions), I-18 is
$34.84\pm3.58$ for primary AHC, $37.22\pm2.48$ for additional AHC, and
$32.62\pm3.06$ for fresh native PPO, following the same ordering as return.

\paragraph{Paired change between budgets.}
For each cohort, the paired change is the same training run and seed's
return at update 15300 minus its return at update 7600. Across three paired
seeds, the mean $\pm$ sample SD is $+8.01\pm1.70$ for primary AHC,
$+5.69\pm2.55$ for additional AHC, and $+5.28\pm0.91$ for fresh native PPO.
Primary AHC gains the most between the two budgets.

\paragraph{Same GPU model.}
The nine D4 intervention runs and all three fresh full-budget native-PPO runs used the same GPU model (Table~\ref{tab:ablation-spread}), giving a hardware-matched comparison. At 2.0054016B PPO rollout transitions, their mean returns are $46.10\pm1.18$ for D4-A, $46.42\pm1.55$ for D4-H, $45.14\pm2.79$ for D4-AH, and $41.91\pm1.01$ for native PPO: all three ablation arms remain 3.2--4.5 points above native PPO. The joint arm's Welch 95\% interval is [$-2.87$, 9.33], and its difference from primary AHC is $-1.56$ [$-8.15$, 5.02] (Table~\ref{tab:return-contrasts}).

\paragraph{Intermediate capability profiles.}
Intermediate-18 is the macro success rate over the official 18-achievement Intermediate set. It decomposes into four mine-focused achievements (M4: mine entry, warrior, archer, bat) and the other 14 (O14): $\text{I-18}=(4\,\text{M4}+14\,\text{O14})/18$. At the full rollout budget, primary AHC and fresh native PPO have M4 rates of 19.86\% and 1.36\%, and O14 rates of 56.28\% and 52.95\%. The 6.69-point Intermediate-18 difference contains approximately 4.11 points from M4 and 2.58 from O14. At update 7600, their O14 rates are 43.96\% and 41.91\%, using the fresh native-PPO 7600 checkpoint (Table~\ref{tab:tier2-skills}).

\paragraph{Seed-dependent onset of deep-floor combat.}
Warrior success appears at different times in different runs. At checkpoint
7600, AHC seeds 0 and 1 defeat a warrior in 0.99\% and 1.05\% of the
episodes that enter the mines, whereas seed 2 does so in 68.50\% (448 of
654). By checkpoint 15300 all three
are at 74.9--79.0\%. Training records agree: the first session whose logged warrior rate exceeds
1\% is session 105, 102, and 40 for seeds 0--2, and checkpoint 7600 closes
session 74. At checkpoint 7600, the cross-seed ordering of return and Intermediate-18
does not match that of conditional warrior success:
seed 2 has the lowest return (36.08) and Intermediate-18 (30.70\%). Similarly, D4-H
seeds 0 and 1 have nearly equal Intermediate-18 (47.85\% and 47.96\%) but warrior
rates given mine entry of 0\% (0 of 3274) and 80.33\% (2250 of 2801).
Within the primary and additional full-AHC cohorts at the two reported checkpoints,
no model defeats a warrior in all ten rollouts of any world; joint-ablation
seed 0 does so on two worlds and has mixed outcomes on 884 of its 1024
worlds (Appendix~\ref{apx:unified}).

\paragraph{Sources of variability.}
The records separate variation in acquiring a policy across training runs from variation in rollout outcomes on fixed worlds. At update 7600, additional AHC seed 4 records 397 warrior successes in 697 mine-entry episodes (56.96\%), while seeds 1 and 5 record 0 of 470 and 2 of 1305 (0.15\%). Runs with the same architecture and curriculum therefore acquire deep-floor combat at very different times.

\paragraph{Full-budget native PPO.}
Three native-PPO runs start from scratch with training seeds 0--2, the original task, student, optimizer, PPO and environment settings, and numerical guards. The only change to these settings is the session-level entropy stopping threshold, set to $-\infty$; the PPO entropy regularization coefficient remains 0.002. All three complete 15,300 accepted updates (2,005,401,600 PPO rollout transitions; Appendix~\ref{apx:ppo-stopped}). Their endpoint returns are 42.35, 40.74, and 42.62 ($41.91\pm1.01$), 0.50 points above the older guard-stopped cohort. At the endpoint, 1,668 of 30,720 episodes enter the mines and none defeats a warrior; seed 2 records one archer and one bat success in different episodes. We evaluate these runs at updates 7600 and 15300.

\paragraph{Additional cohort.}
The three additional full-AHC executions and the additional fixed-difficulty run
were evaluated with the same repeated-world schedule (Table~\ref{tab:unified}). The full-AHC executions
average $46.08\pm1.23$ (45.40, 47.50, and 45.33 for seeds 1, 4, and 5); the
fixed-difficulty run returns 46.20. The deep-floor divergence seen under
one rollout persists: warrior success occurs in 0.00\%, 24.58\%, and 1.82\%
of episodes for seeds 1, 4, and 5, and among them only seed 4 retains the
four-skill boundary; the fixed-difficulty run also retains it (18.80\%
warrior success). Intermediate-18 averages $46.77\pm2.06\%$, while O14 varies only
from 56.13\% to 57.38\%, so the spread is again concentrated in M4.

\paragraph{All-achievement outcomes and return composition.}
The complete 67-achievement profile is shown in
Figures~\ref{fig:all67a}--\ref{fig:all67b}, including negative and zero
differences. Each cell compares
the mean of three training-seed rates with fresh native PPO at update 15300.
Table~\ref{tab:return-decomposition} partitions the observed mean-return
difference using the pinned native reward weights. For primary AHC, 3.615 of
the 4.795-point gap comes from the 18 weight-3 Intermediate achievements;
1.115 comes from weight-1 achievements. In the additional cohort the
corresponding contributions are 2.854 and 1.128 points. The nine weight-8
achievements have no successes in any of these six cohorts, so the deepest
floors remain unsolved. Individual rates also reveal tradeoffs; for example,
wake-up success is lower in both full-AHC cohorts than in fresh PPO.
Appendix~\ref{apx:all-capabilities} gives the calculation.

\begin{table}[htb]
\centering\small
\caption{Native-weight decomposition of the full-budget mean-return gap
relative to fresh native PPO. Contributions are in return points. Each
cohort has three seeds.}
\label{tab:return-decomposition}
\begin{tabular}{lrrrrrr}
\toprule
Cohort & Total & $w=1$ (25) & $w=3$ (18) & $w=5$ (15) & $w=8$ (9) & Residual \\
\midrule
L40S AHC & 4.795 & 1.115 & 3.615 & 0.072 & 0 & $-0.007$ \\
A100 AHC & 4.171 & 1.128 & 2.854 & 0.187 & 0 & 0.002 \\
D4-A & 4.196 & 1.046 & 3.038 & 0.116 & 0 & $-0.004$ \\
D4-H & 4.512 & 0.527 & 3.955 & 0.021 & 0 & 0.009 \\
D4-AH & 3.230 & 0.696 & 2.508 & 0.021 & 0 & 0.004 \\
\bottomrule
\end{tabular}
\end{table}

\subsection{Three-seed component ablations}
\label{sec:ablations}
Each of the three intervention arms has three training runs of 15,300
updates (2,005,401,600 PPO rollout transitions). Table~\ref{tab:unified}
reports their checkpoint-15300 repeated-world results; single-rollout
evaluations are in Appendix~\ref{apx:d4}.

\paragraph{Interventions.}
D4-A freezes each unit's difficulty parameter at its value when that unit
is created. Selection, graduation, parking, and independent variants remain
active.
D4-H removes inherited progress from the signature used to initialize a
practice state. Bank admission, variant-defined prefills, resource and lighting
adjustments, and independent variants remain active.
The fresh D4-AH runs apply both interventions simultaneously while preserving
the student, optimizer, task-selection rules, original-task anchor, bank
admission, variant-defined prefills, and the remaining state operations. Thus, this arm
removes the two tested mechanisms, while the rest of AHC remains active.

\begin{table}[htb]
\centering\small
\setlength{\tabcolsep}{4pt}
\caption{Full-budget component comparisons under the same repeated-world
evaluation. Each arm has three training seeds; return, M4, and O14 show
mean $\pm$ sample SD. Lowest and highest are the observed per-seed return
means, each estimated from 10,240 episodes.}
\label{tab:ablation-spread}
\begin{tabular}{P{3.05cm}crrrrr}
\toprule
Arm & GPU & Return & Lowest & Highest & M4 (\%) & O14 (\%) \\
\midrule
Full AHC, primary & L40S & $46.70\pm0.57$ & 46.05 & 47.10 & $19.86\pm1.47$ & $56.28\pm0.70$ \\
D4-A: fixed difficulty & A6000 & $46.10\pm1.18$ & 45.22 & 47.45 & $15.81\pm6.51$ & $56.06\pm0.77$ \\
D4-H: no progress init. & A6000 & $46.42\pm1.55$ & 44.73 & 47.78 & $18.64\pm9.48$ & $57.43\pm1.90$ \\
D4-AH: joint & A6000 & $45.14\pm2.79$ & 42.68 & 48.17 & $11.51\pm15.30$ & $56.03\pm2.10$ \\
\midrule
Full AHC, additional & A100 & $46.08\pm1.23$ & 45.33 & 47.50 & $11.98\pm11.02$ & $56.71\pm0.63$ \\
\bottomrule
\end{tabular}
\end{table}

\paragraph{Return and training-seed variability.}
Relative to the primary cohort, mean return is lower by 0.60 points for
D4-A, 0.28 for D4-H, and 1.56 for D4-AH. Return SD is 0.57 for primary full
AHC, 1.18 and 1.55 for the two single interventions, and 2.79 for their
joint intervention (Table~\ref{tab:ablation-spread}). The best seeds stay
high (47.10, 47.45, 47.78, and 48.17), whereas the worst fall from 46.05 to
45.22, 44.73, and 42.68: in these runs, removing the mechanisms lowers the
worst seed rather than the best.

\paragraph{Capability allocation and repeated-world variability.}
The joint arm's I-18 is $46.13\pm4.78$\%, with M4
$11.51\pm15.30$\% and O14 $56.03\pm2.10$\%.
Relative to D4-A, its I-18 mean is lower by 0.98 percentage points.
Using $\mathrm{I18}=(4\mathrm{M4}+14\mathrm{O14})/18$, the weighted
M4 term contributes $-0.96$ points and the O14 term $-0.02$.
O14 includes dungeon and orc-combat achievements.
D4-H's I-18 mean (48.81\%) is slightly above the primary cohort's 48.18\%;
removing one component does not uniformly reduce all capabilities.

Joint seed 0 returns 48.17 with M4 29.11\%; seeds 1 and 2 return 44.56
and 42.68 with M4 4.08\% and 1.34\%. The latter runs enter the mines
1667 and 544 times but record only one and two warrior successes,
respectively. Conditional warrior success is 83.69\%, 0.06\%, and
0.37\% across joint seeds 0--2. This extends the separation of access and combat
already seen in D4-H seed 0 (3274 mine entries and no warrior success).
The four-skill boundary is retained in three D4-A runs,
two D4-H runs, and one D4-AH run.
Joint seed 0 still has mixed warrior outcomes on 884 of the 1024 worlds
and success on all ten repeats on two worlds. Removing both tested
mechanisms therefore preserves both training-seed divergence and
within-world stochastic variation.

\subsection{Compute cost}
\label{sec:compute-ledger}
\label{sec:cost}
AHC executes no LLM inference in task proposal, mutation, embedding, training,
or evaluation. Table~\ref{tab:cost} gives the allocation time of the three
primary training jobs, each on one NVIDIA L40S: 53.0147 L40S GPU$\cdot$h in
total, including training-period evaluations.

\begin{table}[htb]
\centering\small
\caption{Allocation time for the three primary AHC training jobs,
including in-job evaluation. Each job used one L40S, so GPU$\cdot$h equals
elapsed hours.}
\label{tab:cost}
\begin{tabular}{P{3.8cm}P{4.4cm}P{5.6cm}}
\toprule
Training seed & Elapsed (hh:mm:ss) & Allocated L40S GPU$\cdot$h \\
\midrule
0 & 17:50:18 & 17.8383 \\
1 & 17:33:14 & 17.5539 \\
2 & 17:37:21 & 17.6225 \\
\midrule
Completed-job total & 53:00:53 & 53.0147 \\
\bottomrule
\end{tabular}
\end{table}

The fresh full-budget native-PPO runs record 23.2864, 23.2814, and 23.3934 RTX A6000 GPU$\cdot$h for seeds 0--2 (69.9612 in total), and the joint D4-AH runs 85.2268 RTX A6000 GPU$\cdot$h. The additional AHC cohort records at least 54.594 A100 GPU$\cdot$h. Historical 14B-teacher curriculum runs averaged 58.9 GPU$\cdot$h. Each AHC run performs 151 harvest operations over 1024 session-ending states (154,624 states); these states come from counted rollouts, so harvesting adds no environment transitions.

\section{Related Work}
\label{sec:related}
\paragraph{Environment design and replay.}
PAIRED formalizes environment design through regret
\citep{dennis2020paired}; PLR and replay-guided adversarial design prioritize
previously encountered levels \citep{jiang2021plr,jiang2021robustplr}.
ACCEL combines replay with level mutation \citep{parkerholder2022accel}.
Refinements of regret objectives and analyses of their approximations
\citep{beukman2024refining,rutherford2024noregrets} motivate care in
interpreting success-based scores. AHC adopts a PLR-style rank/staleness
selector and an inherited learnability term, and organizes
achievement-oriented practice around harvested progress signatures, explicit
difficulty control, and native-task anchoring. Its candidate generation is
engine-informed rather than environment-general.

\paragraph{Start-state curricula.}
Harvested initialization places AHC among methods that begin practice from
states other than the native reset. Reverse curriculum generation expands
start states backward from a goal \citep{florensa2017reverse};
demonstration-based resets move the start progressively earlier along an
expert trajectory \citep{resnick2018backplay,salimans2018montezuma};
Go-Explore returns to archived states before exploring
\citep{ecoffet2021first}; and jump-start RL uses a guide policy to reach the
states from which a learner continues \citep{uchendu2023jsrl}. AHC
constructs practice start states by combining procedurally generated worlds
with signatures harvested from the student's earlier progress, which makes
its deep-floor practice contingent on that progress.

\paragraph{LLM curricula and executable tasks.}
OMNI and OMNI-EPIC use models to propose interesting tasks or executable
environments \citep{zhang2024omni,faldor2025omniepic}; EnvGen and Eurekaverse
adapt training environments with LLM generation
\citep{zala2024envgen,liang2024eurekaverse}. DiCode integrates generated
levels with curriculum learning \citep{mitsides2026dicode}. AHC replaces
that generation route with fixed task enumeration and task-state operators.

\paragraph{Action filtering and empirical measurement.}
Work on valid-action suppression distinguishes fully unmasked policies from
oracle and learned-mask deployment \citep{zabounidis2026suppression}.
AHC belongs to the full-action-support category. Measurement studies stress
run-level variability, reproducible comparisons, and uncertainty in RL
\citep{henderson2018matters,agarwal2021precipice,patterson2024empirical}.
We therefore report per-seed results and capability rates alongside return,
and separate endpoint from tail-window statistics.

\section{Limitations and Conclusion}
\label{sec:discussion}
\paragraph{Limitations.}
AHC relies on domain-specific task rules, a hand-specified dependency graph,
harvested training progress, and modified training dynamics. Its native,
fully unmasked evaluation does not make its training conditions identical
to those of native PPO, and transfer to other environments is untested.

The comparison with native PPO evaluates complete training systems. We did
not run an equally assisted uniform selector or a matched non-LLM
curriculum, such as the PLR, SFL, and DR baselines in the DiCode
implementation \citep{mitsides2026dicode}, and the original-task anchor and
selection rule are not ablated; the controller's added value over simpler
curricula therefore remains open. Primary AHC (L40S), the additional cohort
(A100), and the ablation and native-PPO runs (RTX A6000) were trained on
different GPU models, which limits attribution of differences to individual
components and their interaction. The wider seed spread of the ablations
suggests that harvested progress and difficulty adaptation may make
capability acquisition more consistent; with three seeds per arm, and with
the additional cohort's SD (1.23) slightly above D4-A's (1.18), this remains
a hypothesis.

Each cohort has three training seeds, and the Welch intervals are not
adjusted for multiple comparisons. The primary cohort was designated during
manuscript revision rather than prospectively, which is why we report the
additional cohort separately. Repeated rollouts sharpen estimates on the
1024 evaluation worlds but do not add training seeds. Budget matching covers
accepted PPO rollout transitions only: we did not meter total environment
interaction, and the GPU-hours in \secref{sec:cost} come from different
hardware, so we make no claim about compute efficiency or LLM inference cost.

The diagnosis concerns one environment and one 14B teacher. Graph
comparisons bundle graph and API-reference changes, one Surgical run
contains a level-generation defect (17.6\% of its mob-placement calls place
a passive mob), and the four deep-floor skills in the profile comparison
were chosen after inspecting the runs, so the observed separations are
descriptive. Cross-study comparisons with DiCode and external PPO-GTrXL
results differ in software versions, evaluation protocols, and action
support, and serve only as context.

\paragraph{Conclusion.}
AHC combines progress reuse, difficulty adaptation, bounded selection, and original-task practice without LLM inference. Under fully unmasked native Craftax evaluation, the primary cohort reaches $46.70\pm0.57$ at 2.0054016B PPO rollout transitions, compared with $41.91\pm1.01$ for fresh native PPO-GTrXL, and $38.69\pm2.26$ versus $36.63\pm1.42$ at 0.9961472B; an additional AHC cohort reaches $46.08\pm1.23$ at the full budget. Most of the gain comes from Intermediate achievements, especially the four mine-floor achievements, while the deepest-floor achievements remain unsolved. Single-component ablations stay near the primary mean, and the joint intervention gives $45.14\pm2.79$, with one high-performing seed and two runs that rarely succeed in mine combat. The accompanying diagnosis shows how prerequisite and floor-access rules restrict admissible practice in an LLM curriculum, and why capability profiles should be reported alongside aggregate return.

\subsubsection*{Reproducibility Statement}
The manuscript specifies task construction, controller rules, the shared
PPO/GTrXL configuration, and the frozen native-task evaluation protocol.
Retained materials include source and configuration archives, checkpoint
inventories, complete episode matrices, and result-derivation code; the
appendices give the controller specification, state interface, and per-seed
results.

\subsubsection*{Statement on LLM Usage}
Large language models assisted with manuscript editing and analysis-code
drafting. The authors are responsible for verifying text, calculations,
citations, and experimental provenance. The AHC procedure itself performs
no LLM inference.

\bibliography{references_main10}

@inproceedings{mitsides2026dicode,
  title     = {Dreaming in Code for Curriculum Learning in Open-Ended Worlds},
  author    = {Mitsides, Konstantinos and Faldor, Maxence and Cully, Antoine},
  booktitle = {Proceedings of the 43rd International Conference on Machine Learning},
  year      = {2026},
  url       = {https://proceedings.mlr.press/v306/mitsides26a.html},
  note      = {Anchors quoted here are the camera-ready 8-seed set}
}

@inproceedings{matthews2024craftax,
  title     = {Craftax: A Lightning-Fast Benchmark for Open-Ended Reinforcement Learning},
  author    = {Matthews, Michael and Beukman, Michael and Ellis, Benjamin and Samvelyan, Mikayel and Jackson, Matthew and Coward, Samuel and Foerster, Jakob},
  booktitle = {Proceedings of the 41st International Conference on Machine Learning (ICML)},
  year      = {2024},
  note      = {arXiv:2402.16801}
}

@inproceedings{hafner2022crafter,
  title     = {Benchmarking the Spectrum of Agent Capabilities},
  author    = {Hafner, Danijar},
  booktitle = {International Conference on Learning Representations (ICLR)},
  year      = {2022},
  note      = {arXiv:2109.06780}
}

@inproceedings{parisotto2020gtrxl,
  title     = {Stabilizing Transformers for Reinforcement Learning},
  author    = {Parisotto, Emilio and Song, H. Francis and Rae, Jack W. and Pascanu, Razvan and Gulcehre, Caglar and Jayakumar, Siddhant M. and Jaderberg, Max and Kaufman, Raphael Lopez and Clark, Aidan and Noury, Seb and Botvinick, Matthew M. and Heess, Nicolas and Hadsell, Raia},
  booktitle = {Proceedings of the 37th International Conference on Machine Learning (ICML)},
  year      = {2020},
  note      = {arXiv:1910.06764}
}

@article{schulman2017ppo,
  title   = {Proximal Policy Optimization Algorithms},
  author  = {Schulman, John and Wolski, Filip and Dhariwal, Prafulla and Radford, Alec and Klimov, Oleg},
  journal = {arXiv preprint arXiv:1707.06347},
  year    = {2017}
}

@inproceedings{dennis2020paired,
  title     = {Emergent Complexity and Zero-shot Transfer via Unsupervised Environment Design},
  author    = {Dennis, Michael and Jaques, Natasha and Vinitsky, Eugene and Bayen, Alexandre and Russell, Stuart and Critch, Andrew and Levine, Sergey},
  booktitle = {Advances in Neural Information Processing Systems (NeurIPS)},
  volume    = {33},
  year      = {2020},
  note      = {arXiv:2012.02096}
}

@inproceedings{jiang2021plr,
  title     = {Prioritized Level Replay},
  author    = {Jiang, Minqi and Grefenstette, Edward and Rocktaschel, Tim},
  booktitle = {Proceedings of the 38th International Conference on Machine Learning (ICML)},
  year      = {2021},
  note      = {arXiv:2010.03934}
}

@inproceedings{jiang2021robustplr,
  title     = {Replay-Guided Adversarial Environment Design},
  author    = {Jiang, Minqi and Dennis, Michael and Parker-Holder, Jack and Foerster, Jakob and Grefenstette, Edward and Rocktaschel, Tim},
  booktitle = {Advances in Neural Information Processing Systems (NeurIPS)},
  volume    = {34},
  year      = {2021},
  note      = {arXiv:2110.02439}
}

@inproceedings{parkerholder2022accel,
  title     = {Evolving Curricula with Regret-Based Environment Design},
  author    = {Parker-Holder, Jack and Jiang, Minqi and Dennis, Michael and Samvelyan, Mikayel and Foerster, Jakob and Grefenstette, Edward and Rocktaschel, Tim},
  booktitle = {Proceedings of the 39th International Conference on Machine Learning (ICML)},
  year      = {2022},
  note      = {arXiv:2203.01302}
}

@inproceedings{beukman2024refining,
  title     = {Refining Minimax Regret for Unsupervised Environment Design},
  author    = {Beukman, Michael and Coward, Samuel and Matthews, Michael and Fellows, Mattie and Jiang, Minqi and Dennis, Michael and Foerster, Jakob},
  booktitle = {Proceedings of the 41st International Conference on Machine Learning (ICML)},
  year      = {2024},
  note      = {arXiv:2402.12284}
}

@inproceedings{rutherford2024noregrets,
  title     = {No Regrets: Investigating and Improving Regret Approximations for Curriculum Discovery},
  author    = {Rutherford, Alexander and Beukman, Michael and Willi, Timon and Lacerda, Bruno and Hawes, Nick and Foerster, Jakob},
  booktitle = {Advances in Neural Information Processing Systems (NeurIPS)},
  volume    = {37},
  year      = {2024},
  note      = {arXiv:2408.15099}
}

@inproceedings{zhang2024omni,
  title     = {{OMNI}: Open-Endedness via Models of Human Notions of Interestingness},
  author    = {Zhang, Jenny and Lehman, Joel and Stanley, Kenneth and Clune, Jeff},
  booktitle = {International Conference on Learning Representations (ICLR)},
  year      = {2024},
  note      = {arXiv:2306.01711}
}

@inproceedings{faldor2025omniepic,
  title     = {{OMNI-EPIC}: Open-Endedness via Models of Human Notions of Interestingness with Environments Programmed in Code},
  author    = {Faldor, Maxence and Zhang, Jenny and Cully, Antoine and Clune, Jeff},
  booktitle = {International Conference on Learning Representations (ICLR)},
  year      = {2025},
  note      = {arXiv:2405.15568}
}

@inproceedings{liang2024eurekaverse,
  title     = {Environment Curriculum Generation via Large Language Models},
  author    = {Liang, William and Wang, Sam and Wang, Hung-Ju and Bastani, Osbert and Jayaraman, Dinesh and Ma, Yecheng Jason},
  booktitle = {Proceedings of The 8th Conference on Robot Learning},
  series    = {Proceedings of Machine Learning Research},
  volume    = {270},
  pages     = {433--454},
  year      = {2025},
  publisher = {PMLR},
  url       = {https://proceedings.mlr.press/v270/liang25a.html}
}

@inproceedings{zala2024envgen,
  title     = {EnvGen: Generating and Adapting Environments via LLMs for Training Embodied Agents},
  author    = {Zala, Abhay and Cho, Jaemin and Lin, Han and Yoon, Jaehong and Bansal, Mohit},
  booktitle = {Conference on Language Modeling (COLM)},
  year      = {2024},
  note      = {arXiv:2403.12014}
}

@inproceedings{agarwal2021precipice,
  title     = {Deep Reinforcement Learning at the Edge of the Statistical Precipice},
  author    = {Agarwal, Rishabh and Schwarzer, Max and Castro, Pablo Samuel and Courville, Aaron and Bellemare, Marc G.},
  booktitle = {Advances in Neural Information Processing Systems (NeurIPS)},
  volume    = {34},
  year      = {2021},
  note      = {arXiv:2108.13264}
}

@article{patterson2024empirical,
  title   = {Empirical Design in Reinforcement Learning},
  author  = {Patterson, Andrew and Neumann, Samuel and White, Martha and White, Adam},
  journal = {Journal of Machine Learning Research},
  volume  = {25},
  number  = {318},
  pages   = {1--63},
  year    = {2024},
  note    = {arXiv:2304.01315}
}

@inproceedings{henderson2018matters,
  title     = {Deep Reinforcement Learning that Matters},
  author    = {Henderson, Peter and Islam, Riashat and Bachman, Philip and Pineau, Joelle and Precup, Doina and Meger, David},
  booktitle = {Proceedings of the 32nd AAAI Conference on Artificial Intelligence},
  year      = {2018},
  note      = {arXiv:1709.06560}
}

@article{zabounidis2026suppression,
  title   = {Overcoming Valid Action Suppression in Unmasked Policy Gradient Algorithms},
  author  = {Zabounidis, Renos and Siegelmann, Roy and Qadri, Mohamad and Kim, Woojun and Stepputtis, Simon and Sycara, Katia P.},
  journal = {arXiv preprint arXiv:2603.09090},
  year    = {2026},
  url     = {https://arxiv.org/abs/2603.09090}
}

@misc{hamon2024transformerxl,
  title        = {transformerXL\_PPO\_JAX},
  author       = {Hamon, Gautier},
  year         = {2024},
  howpublished = {Software repository},
  url          = {https://github.com/Reytuag/transformerXL_PPO_JAX},
  note         = {Craftax repository result: 20.6\% of maximum reward at 4 billion timesteps}
}

@inproceedings{florensa2017reverse,
  title     = {Reverse Curriculum Generation for Reinforcement Learning},
  author    = {Florensa, Carlos and Held, David and Wulfmeier, Markus and Zhang, Michael and Abbeel, Pieter},
  booktitle = {Proceedings of the 1st Annual Conference on Robot Learning},
  series    = {Proceedings of Machine Learning Research},
  volume    = {78},
  pages     = {482--495},
  year      = {2017}
}

@article{salimans2018montezuma,
  title   = {Learning {Montezuma's Revenge} from a Single Demonstration},
  author  = {Salimans, Tim and Chen, Richard},
  journal = {arXiv preprint arXiv:1812.03381},
  year    = {2018}
}

@article{resnick2018backplay,
  title   = {Backplay: ``{Man} muss immer umkehren''},
  author  = {Resnick, Cinjon and Raileanu, Roberta and Kapoor, Sanyam and Peysakhovich, Alexander and Cho, Kyunghyun and Bruna, Joan},
  journal = {arXiv preprint arXiv:1807.06919},
  year    = {2018}
}

@article{ecoffet2021first,
  title   = {First return, then explore},
  author  = {Ecoffet, Adrien and Huizinga, Joost and Lehman, Joel and Stanley, Kenneth O. and Clune, Jeff},
  journal = {Nature},
  volume  = {590},
  number  = {7847},
  pages   = {580--586},
  year    = {2021}
}

@inproceedings{uchendu2023jsrl,
  title     = {Jump-Start Reinforcement Learning},
  author    = {Uchendu, Ikechukwu and Xiao, Ted and Lu, Yao and Zhu, Banghua and Yan, Mengyuan and Simon, Jos{\'e}phine and Bennice, Matthew and Fu, Chuyuan and Ma, Cong and Jiao, Jiantao and Levine, Sergey and Hausman, Karol},
  booktitle = {Proceedings of the 40th International Conference on Machine Learning},
  series    = {Proceedings of Machine Learning Research},
  volume    = {202},
  pages     = {34556--34583},
  year      = {2023}
}
\bibliographystyle{tmlr}
\clearpage
\appendix
\section{Supplementary Details}

\subsection{AHC controller specification}
\label{apx:ahc}

Table~\ref{tab:ahc-loop} gives the session procedure. The specification
below follows the executed source and environment settings. Paths are
relative to \texttt{src/dicode/}.

\paragraph{Task universe and labels.}
The twelve recipe families are wood/stone/iron/diamond pickaxes,
wood/stone/iron/diamond swords, iron/diamond armour, arrows, and torches.
Entry targets $f=1,\ldots,8$ map respectively to dungeon, gnomish mines,
sewers, vault, troll mines, fire realm, ice realm, and graveyard. Gate
labels 0, 3, 6, and 8 denote recorded kill-count bins 0--2, 3--5, 6--7,
and at least 8 on the source floor; they are not difficulty values. The
ordinary unit set is the 67-achievement enumeration minus these twelve
recipe and eight entry targets. With merged levels enabled, family and
depth bases use L1/L2 rather than four surgery levels. Family subtiering
is disabled, so every family has only i0 with no subtier prefill. Variant
prefills remain separate. L4 is a family-specific official-reset path,
not floor 4 and not an additional dynamics level. It uses official EMA
for scoring rather than a curriculum completion estimate. Non-axis bases,
including family L2, receive $\sigma=0$; explicit variants may supply their
own softening parameters. The post-fixup target-zero assertion is performed
before the assembled bank is attached to resets.

\paragraph{Statistics and relevance.}
\texttt{anneal/curator.py} updates curriculum EMA whenever $n_u>0$, but
advances difficulty and curriculum mastery counters only when $n_u\geq100$.
Official-anchor EMA also uses coefficient 0.5 and requires 100 observations.
The window is \texttt{scoring\_window\_updates=40}: updates 61--100
of a standard 100-update session, each containing 128 steps per environment.
\texttt{ppo\_tr.py} slices the update axis before flattening;
\texttt{anneal/pcount.py} then masks by task identifier and
\texttt{returned\_episode}. Thus $n_u$ is not the number of environment
steps, episode starts, or episodes over the whole session. The logging
wrapper requires episode length greater than one. Counts are refreshed
per session, while the EMA retains past information. Initialization uses
the same final-40-update slice of its 200 updates. Official feedback is
computed from the \texttt{OriginalTask} anchor in that training slice;
its 100-observation gate applies before updating official EMA. Frozen-world
evaluation does not feed the controller, and an absent reading is treated as
missing rather than as zero success.
\texttt{anneal/prereq.py} computes $R$ from the latest raw official rates,
clipped to $[0,1]$, with native reward weights 1, 3, 5, or 8. Its descendant
graph combines recipes and floor dependencies with hand-specified engine
dependencies; it represents practice routes, not every stochastic shortcut.
For graduated units $R$ is set to 1. Official-reset-level units use smoothed
official success in place of $\hat p_u$ and require an available reading.
Other units retain curriculum success as their score input.

\paragraph{Candidate exceptions.}
Family units and the easiest depth levels bypass neighbor unlocking; variants
inherit their parent's checks. Already activated units remain unlocked, and
a missing easier neighbor also permits entry. The chain check passes when
no equipment-derived prerequisite is present or the entire official-reading
map is empty; otherwise an individually missing prerequisite blocks entry.
The frontier admits at most one floor beyond the deepest qualifying floor.
Equivalent variants are folded before selection. The active controller has
no hard depth-share quota: the configured 0.45 depth fraction is telemetry,
not an enforced cap. \texttt{anneal/plr.py} supplies normalized inverse-rank
weights and a 0.30 staleness mixture. Fresh-unit priority can consume all
15 seats, so positive sampling weight does not guarantee a seat each session.

\paragraph{What is annealed.}
\texttt{anneal/surgery.py} implements
$x_j(\sigma)=x_j^{\rm native}+\sigma(x_j^{\rm easy}-x_j^{\rm native})$,
with rounding for the integer trigger distance. In the order needs depletion,
health recovery, passive/melee/ranged spawning, mob health, mob damage, and
melee trigger distance, the native vector is
$(1,1,1,1,1,1,1,10)$ and the easy vector is
$(0.8,2,1,0.5,0.5,0.5,0.5,5)$.
Non-family softened units initialize at $\sigma=1$. The three-reading
hardening window is cleared after a difficulty change. A low current reading
takes precedence over hardening. This schedule changes training dynamics;
held-out evaluation always uses the native environment.

\paragraph{Graduation and parking.}
Three consecutive accepted official EMA readings at or above 0.75 graduate
a unit, even if it has not been selected. A fallback route also graduates a
unit with a curriculum mastery counter of three and official EMA at least
0.75, provided it is no longer softened. The curriculum counter increments at
$\hat p\geq0.75$, resets below 0.65, and is retained between those values;
it is therefore a hysteresis counter, not necessarily three consecutive
above-threshold sessions. Graduated depth units are terminally parked.
A nongraduated softened unit at $\sigma\leq10^{-6}$ with the same
three-count condition is also terminally parked (``arrival'').

The active selector excludes terminal parking, marked \texttt{parked\_at=-2},
and graduated units with official EMA at least 0.95. Ordinary low-success or
easy-task parking flags are not themselves exclusions in this selector.
Graduated units whose stored official EMA is below 0.45 for three controller
updates lose graduation and have terminal parking cleared. This regression rule
requires graduation; it does not guarantee reactivation of an arrival-parked,
nongraduated unit. Ordinary parking is reconsidered after ten sessions.

\paragraph{Allocation and state supply.}
\texttt{training.py} doubles weights of units carrying L1 identifiers.
An L2 unit that reaches $\sigma=0$ does not receive this L1 multiplier.
For each
floor $f$, let $L_f$ be the number of ever-active softened depth units and
$A_f$ the number neither parked nor graduated. If $0<A_f<L_f$, selected
members of that surviving set receive multiplier $L_f/A_f$; otherwise the
multiplier is 1. Counts come from controller state, not just the current seat
list. All task weights, including the unchanged anchor weight, are then
globally normalized.
\texttt{anneal/bank.py} bounds each unit bank at 256 entries. In the
active signature path, exact signatures are deduplicated and oldest
entries are evicted at capacity; sampling first covers represented floors
before drawing remaining entries. The grouping key is the signature's
floor, not the generated world's identity. \texttt{anneal/worldpool.py} supplies 256 worlds per floor
across six floors. The assembler preserves harvested inventory and
equipment signatures and applies configured ore and lighting operations.
No extra mob-supply or fountain operation is enabled. Task variants are
created every two sessions, at most three per event and four per goal.
\texttt{ANN\_SIL=0}: harvested progress affects initial states, not an
auxiliary self-imitation loss.

\subsection{Harvested-state and training-task interface}
\label{apx:state-interface}
This specification follows the active signature path in the pinned source,
\texttt{anneal/signature.py}, \texttt{bank.py}, \texttt{frontier.py},
\texttt{spawn.py}, and \texttt{engine\_ctx.py}. Inherited signatures are
training-state information; they are not held-out worlds or saved policy
trajectories. For a flattened session-ending state $x$, define
$S(x)=(f,g,i,a,e)$ as follows.

\begin{table}[H]
\centering\small
\caption{Stored signature fields and realization semantics. Clamps limit
stored quantities, not fixed amounts granted to the student.}
\label{tab:signature-fields}
\begin{tabular}{P{2.1cm}P{6.4cm}P{5.3cm}}
\toprule
Part & Fields stored & Bounds / assembly \\
\midrule
$f$ & \texttt{player\_level} & Selects floor-indexed pool; variant floors are not overridden. \\
$g$ & Full \texttt{monsters\_killed} vector & Copy compatible vector prefix; retain per-floor gate progress. \\
$i$ & Wood, stone, coal, iron, diamond, sapling; pickaxe and sword tiers & Material caps $(3,2,3,3,3,9)$; tool tiers copied as recorded. \\
$a$ & Four \texttt{inventory.armour} slots & Copy recorded values; no fixed armour grant. \\
$e$ & Torches, arrows, ruby, sapphire, bow, books, potion slots & Caps 12 for torches/arrows, 9 for gems, 3 per potion slot; bow/books copied. \\
Excluded & Map, position, timestep, health/needs, actors, historical achievement bits & Never copy these from the harvested state; derive already-earned achievement flags. \\
\bottomrule
\end{tabular}
\end{table}
For materials, the cap is the maximum single-recipe need including station
construction (table: 2 wood; furnace: 1 stone). The twelve-recipe table is
manually transcribed from engine crafting rules. Sapling has no such recipe
need and uses the implementation fallback cap of 9. Armour/equipment
domains are enabled in the retained configuration. Exact signatures are
deduplicated within each 256-entry FIFO bank; sampling prioritizes represented
floors. No source-world identity or historical achievement vector is stored.

\paragraph{Assembly order.}
For each selected unit $u$, the implementation performs these operations:
\begin{enumerate}\itemsep2pt
\item Draw a signature from its bank (or the configured parent bank for a
variant); reject missing signatures or unavailable floor pools.
\item Choose a generated world at the signature floor. World generation
sets the starting floor before construction, so position and actors come
from that world's ordinary floor-start state.
\item Copy $g,i,a,e$ into the generated state with compatible engine dtypes.
For each prefill scalar $v$, set inventory to $\max(\text{recorded},v)$.
The resolved \texttt{ANN\_SUBTIER=0} leaves only family sublevel $i0$,
whose prefill is empty. Only variant dictionaries supply extra inventory
prefills; harvested inventory is inherited separately.
\item Apply configured ore supply and lighting, followed by the explicit
variant soft specification or the unit's current difficulty setting. These
are separate assistance operators, not fields inferred from the signature.
With \texttt{ANN\_SIGMA=1}, every selected nonvariant unit has a
difficulty specification; non-axis units use $\sigma=0$, the native
parameters, including non-axis L2 units. Variant floors are not overridden.
\item Recompute inventory achievements using the engine; mark already-entered
floors according to the starting level. Assert that the target achievement
remains false, then attach the task identifier and construct the observation.
The reset and auto-reset hooks perform this assembly; the original-task
anchor retains ordinary resets. A selected unit assigned official practice,
or with zero assembled states, also uses an ordinary world reset while
retaining its own task goal and training reward contract.
\end{enumerate}
\begin{table}[H]
\centering\small
\caption{State-assembly pseudocode for the active signature path. All unspecified
state fields come from the generated world until an explicit task operator
changes them. The original-task anchor bypasses curriculum assembly.}
\label{tab:state-pseudocode}
\begin{tabular}{rP{13.6cm}}
\toprule
Step & Operation \\
\midrule
1 & For selected unit $u$, obtain entries from its bank or variant-parent bank.
Official-practice units obtain no entries. Skip entries without a signature. \\
2 & Set $S=(f,g,i,a,e)$; for D4-H (and D4-AH), replace it by $(0,(),(),(),())$.
Set $f^*=f$; variant floors remain unspecified in the resolved configuration. \\
3 & Draw $W$ from the official pool at $f^*$; skip if unavailable.
Clone $W$. Its map, actors, position, timestep and vitals are not inherited. \\
4 & Copy each nonempty signature field $g,i,a,e$, preserving engine dtypes.
Copy array prefixes; merge variant scalar prefill as $\max(\text{signature value},v)$,
with omitted signature scalars defaulting to zero. \\
5 & Ensure goal-recipe ore supply, then configured lighting. Apply explicit
variant softening, otherwise the unit's sigma specification. Non-axis
units have $\sigma=0$ (native). \\
6 & Recompute inventory-derived achievement bits; mark entered floors from
$f^*$. Require the goal bit to remain false. Attach $u$ and build observation. \\
7 & Use the assembled states at reset/auto-reset. If none were assembled,
use an ordinary reset with $u$'s goal and training reward contract. \\
\bottomrule
\end{tabular}
\end{table}

Thus prefilling an item can pre-mark its acquisition achievement and removes
the corresponding later acquisition reward. It does not grant a completion
reward at reset. Empty signature slots mean that no harvested value is copied;
the generated world's default is retained. A recorded zero, in contrast, is a
value to copy. Array fields copy the compatible prefix and leave any remaining
generated entries unchanged.

\paragraph{Executable ordering and D4-H semantics.}
Table~\ref{tab:state-pseudocode} summarizes the implementation order.
D4-H replaces $S$ with $(0,(),(),(),())$ before selecting a world: the five
slots remain floor, gates, inventory, armour, and equipment. Its default floor
is therefore zero, since variant floors are not overridden. The four
empty tuples suppress inherited gate/inventory/armour/equipment writes;
they do not assign zero arrays or erase the bank. Variant prefills, ore and lighting
operations, difficulty operations, and task-specific rewards remain active.
This intervention removes inherited starting-floor progress as well as
inherited possessions, while the other curriculum operators remain. An empty
signature is distinct from an unavailable
signature, which is skipped.

\paragraph{Training task contract.}
A curriculum task subclasses \texttt{OriginalTask}, with a singleton set
of relevant achievements and no hand-marked completed achievements. Its
67-dimensional indicator activates that goal. It succeeds when that goal
becomes true and terminates on success, death, the configured timestep cap,
or boss defeat. The original-task indicator activates all 67 goals.
The active non-\texttt{reward} training mode uses
\texttt{MultiTaskMiniCraftaxEnv}, whose base step reward is
\[
r_t^{\rm native}=\sum_{a=1}^{67}w_a(A_{t+1,a}-A_{t,a})
                 +0.1(h_{t+1}-h_t).
\]
Training uses $r_t^{\rm train}=r_t^{\rm native}
+20\,\mathbf{1}\{\text{task success}\}$; the constant completion bonus is
training assistance. Held-out evaluation uses native rewards, ordinary
resets, and no synthetic-task completion bonus.

\subsection{Complete capability profile and threshold sensitivity}
\label{apx:all-capabilities}
The all-67 analysis reads 18 archived final-checkpoint episode matrices
(184,320 episodes): three seeds each for L40S AHC, additional A100 AHC,
D4-A, D4-H, D4-AH, and fresh native PPO. Each matrix contains the full
1024-world $\times$ 10-repeat grid. Means are computed within each seed and
then equally across the three seeds. Ordering follows the official
Craftax 1.4.5 achievement enum, including its stored spelling
\texttt{defeat\_orc\_solider} (displayed as ``soldier'').

\paragraph{Native-weight return calculation.}
The pinned engine assigns weights 1, 3, 5, and 8 to 25, 18, 15, and 9
achievements, respectively. For a native-reset evaluation, compute
$C_w=\sum_{a:w_a=w}w_a\bar p_a$ and
$\epsilon=\bar R-\sum_w C_w$ from recorded returns and achievement rates.
Table~\ref{tab:return-decomposition} reports cohort differences in these
terms. The residual includes the native health term and numerical
remainder. The I-18/M4/O14 macros are unweighted rates, distinct from these
return contributions.

\paragraph{Threshold sensitivity.}
Multiplying all four historical thresholds by
$\alpha\in\{0.50,0.75,1.00,1.25,1.50\}$ leaves the final-endpoint counts
unchanged: L40S AHC 3/3, A100 AHC 1/3, D4-A 3/3, D4-H 2/3, D4-AH 1/3,
and fresh native PPO 0/3.

\begin{figure}[p]
\centering
\includegraphics[width=.95\linewidth]{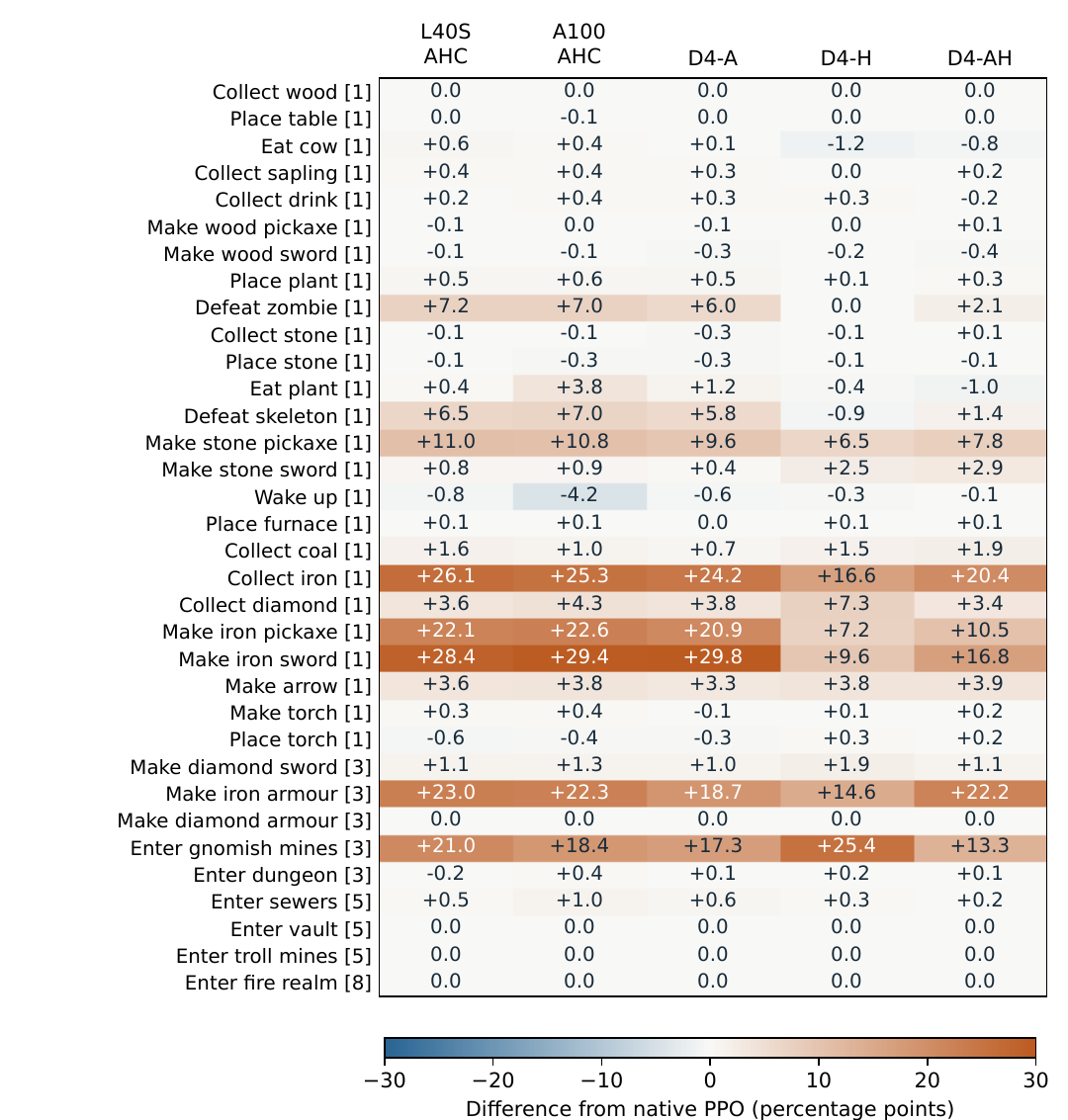}
\caption{All-67 capability differences, part 1: native achievement IDs
0--33. Cells are cohort mean rates minus fresh native PPO in percentage
points at update 15300; brackets give native reward weights. All cells use
three seed means, each based on 1024 fixed worlds and ten repeats.
Zero after rounding need not mean exact equality.}
\label{fig:all67a}
\end{figure}
\begin{figure}[p]
\centering
\includegraphics[width=.95\linewidth]{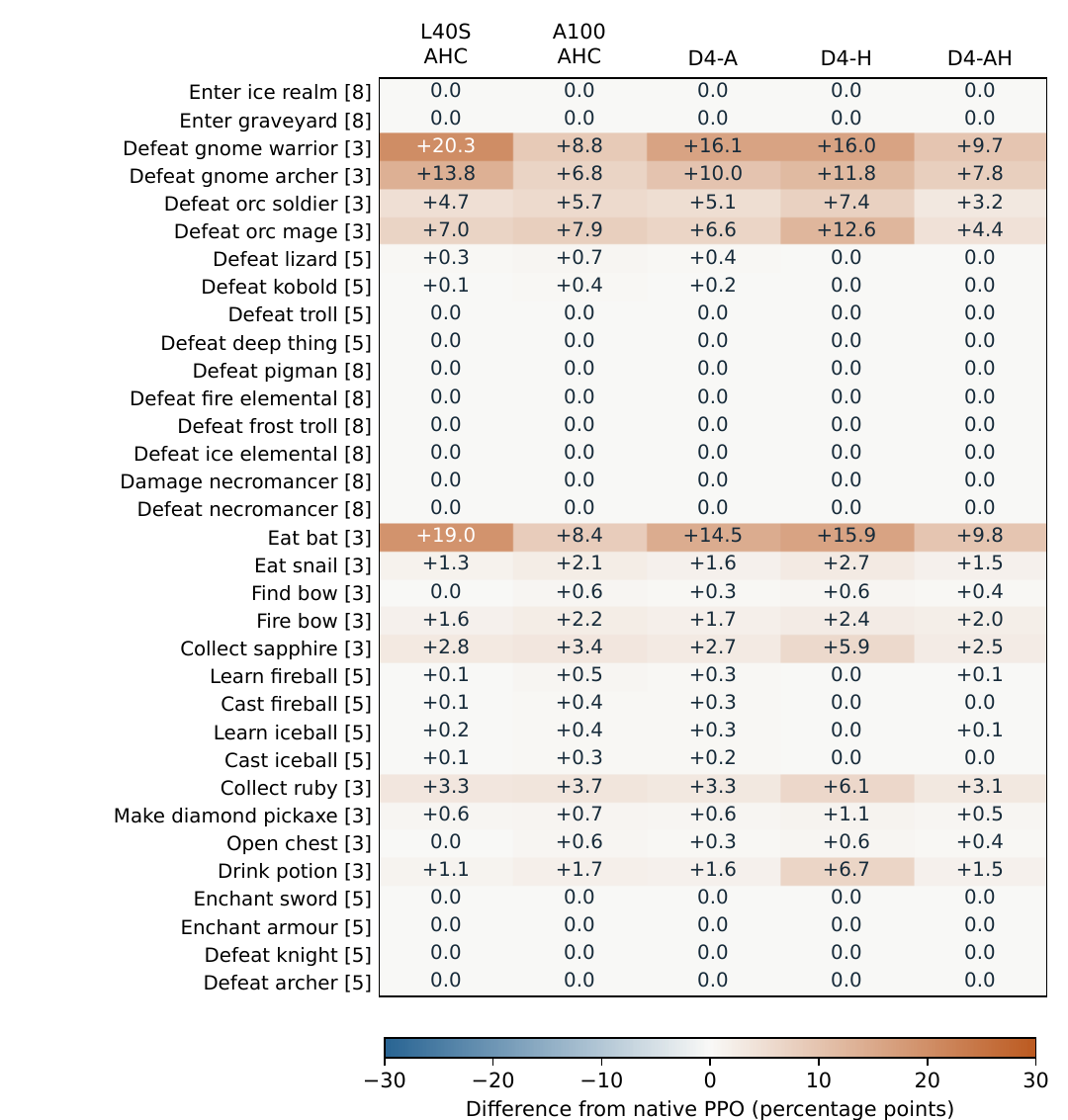}
\caption{All-67 capability differences, part 2: native achievement IDs
34--66, with the same scale and definitions as Figure~\ref{fig:all67a}.
The nine weight-8 achievements have zero successes in every cohort;
zero differences alone would not distinguish shared failure from shared
mastery.}
\label{fig:all67b}
\end{figure}

\subsection{Available curriculum-selection and lifecycle records}
\label{apx:task-trajectories}
The fresh joint arm retains all 151 session task lists and configured
selection probabilities for each of its three seeds. Its first session
is anchor-only; the remaining 150 sessions assign the anchor a configured
share of 0.20. The session-average configured anchor share is therefore
0.2053 in each run.
The retained per-task trajectories also include lifecycle events in the
training log. Ordinary parking events number 37, 39, and 55 for seeds
0--2, and revisit events number 32, 35, and 51; parked tasks are therefore
revisited in this arm.

\begin{table}[H]
\centering\small
\caption{Examples of observed task selection in the joint arm. Counts
refer to sessions with positive configured task probability; first/last
are session indices, not a continuous active interval.}
\label{tab:task-exposure}
\begin{tabular}{lrrr}
\toprule
Task UID & Seed 0 & Seed 1 & Seed 2 \\
\midrule
\texttt{fam:iron\_pickaxe|L1|i0}: selected sessions & 24 & 29 & 34 \\
First / last selection & 2 / 144 & 2 / 148 & 2 / 148 \\
\texttt{depth:f1g8|L2}: selected sessions & 7 & 12 & 32 \\
First / last selection & 2 / 21 & 2 / 23 & 2 / 54 \\
\bottomrule
\end{tabular}
\end{table}
For example, joint seed 1 logs parking of
\texttt{fam:iron\_armour|L2|i0} at session 37 and its revisit at session 47.

The primary full-AHC records additionally retain session task lists,
unit timelines, and controller snapshots for all three training seeds.
Each export covers 151 sessions and 147 configured base units plus the
observed variants (Table~\ref{tab:primary-task-records}). The exporter
records the first logged exit per unit.
Success-rate and difficulty fields are available at the retained snapshot
sessions 8, 18, $\ldots$, 148, and 151, rather than at every session.

\begin{table}[H]
\centering\small
\caption{Scope of recovered primary-AHC task records. Units include 147
base units and the listed variants. ``Selected'' counts distinct units with
at least one logged seat assignment; ``First exit'' counts distinct units
with a first recorded graduation or parking event.}
\label{tab:primary-task-records}
\begin{tabular}{lrrrr}
\toprule
Training seed & Units in export & Variants & Selected & First exit \\
\midrule
0 & 191 & 44 & 132 & 82 \\
1 & 203 & 56 & 145 & 84 \\
2 & 200 & 53 & 144 & 82 \\
\bottomrule
\end{tabular}
\end{table}

\subsection{Evaluated checkpoints}
\label{apx:binding-status}
Table~\ref{tab:retained-inventory} lists the 37 evaluated checkpoints.
\begin{table}[H]
\centering\small
\caption{Repeated-world checkpoint inventory. Every case has
10,240 episode records.}
\label{tab:retained-inventory}
\begin{tabular}{lrrl}
\toprule
Cohort & Runs & Cases & Retained updates \\
\midrule
Primary L40S AHC & 3 & 6 & 7600, 15300 \\
Additional A100 AHC & 3 & 6 & 7600, 15300 \\
Fresh native PPO & 3 & 6 & 7600, 15300 \\
D4-A / D4-H / D4-AH & 3 each & 9 & 15300 \\
Historical guard-stopped PPO & 3 & 9 & 7000, 8000, last accepted endpoint \\
Additional A100 D4-A & 1 & 1 & 15300 \\
\midrule
Total & & 37 & 378,880 evaluated episodes \\
\bottomrule
\end{tabular}
\end{table}
The historical guard-on PPO endpoints stopped at different partial budgets
and provide historical context; the fixed-budget baseline uses the three
completed fresh guard-off PPO runs.

\subsection{Primary and additional AHC cohorts}
\label{apx:ahc-repro}

The primary AHC cohort comprises three compute-cluster jobs, one for each
training seed 0--2. Its
common source, resolved configurations, and complete training logs define
the run set used in Table~\ref{tab:mechanical-endpoint}.

The additional AHC cohort uses seeds 1, 4, and 5 and runs the same code:
its upload archive has the same SHA-256 as the primary source package, all
242 code, configuration, and script files at the execution site match the
archive, the recorded entrypoints and configurations match the launch
script and Hydra configurations, and the 129 pinned dependencies match.
Seed 4 resumed for its final three sessions from checkpoint 15000 with the
saved network, optimizer, curriculum bank, and session count; its controller
RNG and PPO PRNG key were reinitialized from the seed. Repeated-world results
are given in Appendix~\ref{apx:unified}.

\begin{table}[H]
\centering\small
\setlength{\tabcolsep}{4pt}
\caption{Deep-floor success rates (\%) for the six L40S primary and A100
additional runs at checkpoint 15300 under the original single-rollout
protocol. The Remainder column
subtracts the four achievements' contribution (three points times each
success rate) from the return.}
\label{tab:cohort-skills}
\begin{tabular}{llrrrrrr}
\toprule
Training GPU & Seed & Return & Mine entry & Warrior & Archer & Bat & Remainder \\
\midrule
L40S & 0 & 46.861 & 28.52 & 22.36 & 14.45 & 18.95 & 44.33 \\
L40S & 1 & 47.268 & 28.03 & 21.78 & 13.28 & 19.73 & 44.78 \\
L40S & 2 & 46.134 & 23.73 & 18.16 & 14.94 & 18.95 & 43.86 \\
A100 & 1 & 45.205 & 21.78 & 0.00 & 0.00 & 0.00 & 44.55 \\
A100 & 4 & 47.269 & 29.00 & 23.54 & 18.07 & 22.36 & 44.48 \\
A100 & 5 & 44.951 & 19.53 & 1.66 & 0.29 & 0.88 & 44.28 \\
\bottomrule
\end{tabular}
\end{table}

\subsection{Evaluation details and common-endpoint inputs}
\label{apx:measure}
The evaluation helper \nolinkurl{eval_checkpoints.py} calls
\nolinkurl{run_session_evaluation} and then
\nolinkurl{craftax_evaluation.main}. It reconstructs
\texttt{jax.random.PRNGKey(seed)} inside the checkpoint loop. Evaluation
seed 0 determines the 1024 evaluation worlds, which are distinct from the
fixed world pool used for AHC training. The implementation's \texttt{one\_hot}
conditioning is a 67-dimensional achievement indicator, not a
1024-dimensional language embedding. AHC uses \texttt{BatchEnvWrapper}
and first-terminal accumulation; the optimistic-reset setting of 16 is
inactive on that path.

For Table~\ref{tab:local-endpoint}, take the \texttt{15300} entry from each
of the six \texttt{base15\_s*\_eval.json} and
\texttt{longStack\_s*\_eval.json} files in the
\texttt{eval\_summaries} directory. Use \texttt{mean\_return} and the four
\texttt{skills} fields shown below. The three AHC inputs are the primary
final-endpoint JSONs in its retained source package; the seed-0 repeat is
excluded. Aggregate exactly one row per training execution with the sample
SD denominator $n-1$.

\begin{table}[H]
\centering\scriptsize
\setlength{\tabcolsep}{4pt}
\caption{Inputs to the common-checkpoint comparison. These are single
checkpoint-15300 values, not tail-window means. Skill rates are percentages;
six decimal places are shown and JSON precision is retained for aggregation.}
\label{tab:point-inputs}
\begin{tabular}{llrrrrr}
\toprule
System & Training seed & Return & Mine entry & Warrior & Archer & Bat \\
\midrule
AHC & 0 & 46.860676 & 28.515625 & 22.363281 & 14.453125 & 18.945312 \\
AHC & 1 & 47.267670 & 28.027344 & 21.777344 & 13.281250 & 19.726562 \\
AHC & 2 & 46.134350 & 23.730469 & 18.164062 & 14.941406 & 18.945312 \\
base15 & 1 & 46.023933 & 24.804688 & 21.093750 & 14.550781 & 20.507812 \\
base15 & 2 & 42.548748 & 13.964844 & 11.523438 & 7.714844 & 10.644531 \\
base15 & 3 & 44.954964 & 21.191406 & 17.382812 & 7.617188 & 14.062500 \\
longStack & 1 & 45.015549 & 12.402344 & 6.738281 & 0.195312 & 1.464844 \\
longStack & 2 & 43.544601 & 8.007812 & 0.000000 & 0.000000 & 0.976562 \\
longStack & 3 & 45.205635 & 14.941406 & 0.000000 & 0.000000 & 0.000000 \\
\bottomrule
\end{tabular}
\end{table}

\subsection{Repeated-world evaluation design}
\label{apx:d1}
The repeated-world evaluator extends the single-rollout evaluator with
per-episode accumulators and a repeated-play RNG loop. Each repeat reconstructs the same reset batch from evaluation seed
0. Repeat 0 retains the original play key; repeats $r=1,\ldots,9$ use
\texttt{jax.random.fold\_in(play0, r)} for the stream subsequently split for
action sampling and environment transitions. Training seeds identify models,
not evaluation worlds. The 67-dimensional conditioning and unfiltered
43-action policy remain unchanged, and curriculum evaluation is disabled.

Each model has exactly one record for every $(w,r)$ pair in
$\{0,\ldots,1023\}\times\{0,\ldots,9\}$. The 85, 171, and 81 cap-length
episodes for seeds 0--2 are included.

Let $R_{swr}$ be the return for training seed $s$, world $w$, and repeat $r$.
The model mean is $\bar R_s=(1024\cdot10)^{-1}\sum_{w,r}R_{swr}$.
The paper reports the mean and sample SD of the three $\bar R_s$ values.
For within-model world uncertainty, average the ten returns for each world
and divide their sample SD across 1024 worlds by $\sqrt{1024}$. This yields
SEs $(0.189428,0.189670,0.198209)$; these are separate from the sample SD
across training seeds.

The expanded per-model means are $(46.944658,47.104865,46.051475)$, giving
$46.700333\pm0.567608$. Repeat 0 reproduces the earlier single-rollout
evaluations for seeds 0--2, including all 67 skill rates.

\subsection{Repeated-world evaluation of all checkpoints}
\label{apx:unified}
The repeated-world evaluator is applied to 37 checkpoints, 378,880 episodes in total (Table~\ref{tab:retained-inventory}).
Every case has a complete $1024\times10$ repeat-by-world matrix, finite returns, 67 binary achievement fields, native resets, an 8192-step cap, and all 43 native actions available. Evaluation freezes the policy and disables curriculum annealing. Conditioning uses the 67-dimensional all-ones native-task achievement indicator. Repeats change a shared stream for action sampling and environment transitions. Table~\ref{tab:unified-seeds} reports all 37 cases.
Native-PPO checkpoints at updates 7000, 8000, and the guard-stopped endpoints come from one historical three-run set; fresh guard-off checkpoints at updates 7600 and 15300 come from a separate three-run set. The native branch trains only on the original task and does not construct a curriculum or invoke LLM services.

\begin{table}[p]
\centering\scriptsize
\setlength{\tabcolsep}{3pt}
\caption{Per-seed repeated-world results (10,240 episodes per model).
Return SE is world-clustered within each run, not a training-seed SD.
I-18, M4, and O14 are macro-averages (\%) over the 18 official Intermediate
achievements, the four M4 achievements, and the other 14. Mine n and W n
(\%entry) are episode counts; the parenthetical value is the warrior success
percentage among mine-entry episodes. L40S and A100 AHC denote distinct
training cohorts. Fresh native PPO is separate from the historical guard-on
PPO runs.}
\label{tab:unified-seeds}
\begin{tabular}{P{2.55cm}rrrrrrrrr}
\toprule
Model & Seed & Update & Return & SE & I-18 & M4 & O14 & Mine n & W n (\%entry) \\
\midrule
L40S AHC & 0 & 15300 & 46.945 & 0.189 & 48.70 & 20.62 & 56.72 & 2896 & 2202 (76.0) \\
L40S AHC & 1 & 15300 & 47.105 & 0.190 & 48.67 & 20.79 & 56.64 & 2816 & 2226 (79.0) \\
L40S AHC & 2 & 15300 & 46.051 & 0.198 & 47.17 & 18.16 & 55.46 & 2393 & 1793 (74.9) \\
\addlinespace
L40S AHC & 0 & 7600 & 39.924 & 0.187 & 36.92 & 2.52 & 46.74 & 1006 & 10 (1.0) \\
L40S AHC & 1 & 7600 & 40.065 & 0.190 & 36.89 & 1.66 & 46.96 & 666 & 7 (1.1) \\
L40S AHC & 2 & 7600 & 36.078 & 0.237 & 30.70 & 4.54 & 38.18 & 654 & 448 (68.5) \\
\addlinespace
D4-A & 0 & 15300 & 47.446 & 0.204 & 49.27 & 22.62 & 56.88 & 2844 & 2300 (80.9) \\
D4-A & 1 & 15300 & 45.643 & 0.189 & 46.43 & 15.16 & 55.37 & 2145 & 1594 (74.3) \\
D4-A & 2 & 15300 & 45.215 & 0.177 & 45.64 & 9.66 & 55.92 & 1989 & 1037 (52.1) \\
\addlinespace
D4-H & 0 & 15300 & 46.740 & 0.176 & 47.85 & 8.00 & 59.24 & 3274 & 0 (0.0) \\
D4-H & 1 & 15300 & 44.734 & 0.163 & 47.96 & 21.73 & 55.45 & 2801 & 2250 (80.3) \\
D4-H & 2 & 15300 & 47.778 & 0.208 & 50.63 & 26.18 & 57.61 & 3383 & 2663 (78.7) \\
\addlinespace
D4-AH & 0 & 15300 & 48.171 & 0.194 & 51.40 & 29.11 & 57.76 & 3543 & 2965 (83.7) \\
D4-AH & 1 & 15300 & 44.559 & 0.154 & 44.94 & 4.08 & 56.61 & 1667 & 1 (0.1) \\
D4-AH & 2 & 15300 & 42.676 & 0.146 & 42.06 & 1.34 & 53.70 & 544 & 2 (0.4) \\
\addlinespace
Native PPO (guard-on) & 0 & 14600 & 42.086 & 0.159 & 41.84 & 1.82 & 53.27 & 744 & 0 (0.0) \\
Native PPO (guard-on) & 1 & 15000 & 40.249 & 0.154 & 38.78 & 0.57 & 49.70 & 233 & 0 (0.0) \\
Native PPO (guard-on) & 2 & 13600 & 41.867 & 0.139 & 41.76 & 1.17 & 53.36 & 476 & 0 (0.0) \\
\addlinespace
Native PPO (guard-on) & 0 & 7000 & 38.164 & 0.176 & 35.53 & 0.21 & 45.62 & 86 & 1 (1.2) \\
Native PPO (guard-on) & 1 & 7000 & 33.522 & 0.198 & 26.94 & 0.03 & 34.63 & 13 & 0 (0.0) \\
Native PPO (guard-on) & 2 & 7000 & 37.036 & 0.202 & 33.12 & 0.11 & 42.55 & 46 & 0 (0.0) \\
\addlinespace
Native PPO (guard-on) & 0 & 8000 & 38.260 & 0.165 & 36.34 & 0.27 & 46.64 & 109 & 0 (0.0) \\
Native PPO (guard-on) & 1 & 8000 & 37.040 & 0.181 & 33.33 & 0.10 & 42.83 & 38 & 0 (0.0) \\
Native PPO (guard-on) & 2 & 8000 & 39.396 & 0.170 & 36.79 & 0.20 & 47.24 & 80 & 0 (0.0) \\
\addlinespace
A100 AHC & 1 & 15300 & 45.404 & 0.160 & 45.81 & 5.32 & 57.38 & 2177 & 0 (0.0) \\
A100 AHC & 4 & 15300 & 47.495 & 0.207 & 49.14 & 24.70 & 56.13 & 3052 & 2517 (82.5) \\
A100 AHC & 5 & 15300 & 45.330 & 0.163 & 45.36 & 5.91 & 56.63 & 2106 & 186 (8.8) \\
D4-A (A100) & 1 & 15300 & 46.199 & 0.191 & 47.83 & 18.06 & 56.33 & 2545 & 1925 (75.6) \\
\addlinespace
Fresh native PPO & 0 & 15300 & 42.352 & 0.157 & 42.08 & 2.08 & 53.50 & 851 & 0 (0.0) \\
Fresh native PPO & 1 & 15300 & 40.745 & 0.154 & 39.66 & 0.67 & 50.80 & 275 & 0 (0.0) \\
Fresh native PPO & 2 & 15300 & 42.619 & 0.140 & 42.73 & 1.33 & 54.55 & 542 & 0 (0.0) \\
A100 AHC & 1 & 7600 & 39.392 & 0.179 & 35.57 & 1.15 & 45.40 & 470 & 0 (0.0) \\
A100 AHC & 4 & 7600 & 39.431 & 0.199 & 36.03 & 3.81 & 45.24 & 697 & 397 (57.0) \\
A100 AHC & 5 & 7600 & 42.330 & 0.192 & 40.07 & 3.20 & 50.61 & 1305 & 2 (0.2) \\
\addlinespace
Fresh native PPO & 0 & 7600 & 38.077 & 0.181 & 35.79 & 0.22 & 45.95 & 89 & 0 (0.0) \\
Fresh native PPO & 1 & 7600 & 35.239 & 0.190 & 29.67 & 0.09 & 38.12 & 35 & 0 (0.0) \\
Fresh native PPO & 2 & 7600 & 36.566 & 0.194 & 32.41 & 0.06 & 41.66 & 23 & 0 (0.0) \\
\bottomrule
\end{tabular}
\end{table}

For warrior success, let $K$ count the rollouts, out of ten, in which a
world yields a success. None of the first 34 cases has a world with $K=10$;
joint D4-AH seed 0 has two such worlds, 884 mixed worlds, and 138 with
$K=0$. Joint seeds 1 and 2 have 1023 and 1022 worlds with $K=0$,
respectively. At checkpoint
15300 the three AHC seeds have 756--800 worlds with mixed outcomes
($1\leq K\leq9$) and 224--268 with $K=0$. At checkpoint 7600, seeds 0 and 1
have $K=0$ on 1015 and 1017 worlds, whereas seed 2 has 297 mixed worlds.
D4-H seed 0 has $K=0$ on all 1024 worlds, against 814 and 862 mixed worlds
for seeds 1 and 2. Every world has $K=0$ at the native-PPO endpoints and at
update 8000; update 7000 has one mixed world (seed 0). Among the A100
executions, warrior $K=0$ holds on 1024, 178, and 887 worlds for seeds 1, 4,
and 5. The full-budget native-PPO runs have $K=0$ for warrior on every world
and for mine entry on 535, 814, and 670 worlds.

An achievement-order proxy records whether a sword-crafting and an
armour-crafting achievement (iron or diamond) both occurred before the first
mine entry. D4-H seed 0 satisfies it in 1170 of 3274 entries yet records no
warrior success; at checkpoint 7600, AHC seed 2 records 405 of its 448
successes in entries that do not satisfy it. A100 seed 1 likewise satisfies
it in 699 of 2177 entries without a warrior success, and the full-budget
native-PPO runs satisfy it in 246 of 1,668 entries with no warrior success
either way. Crafting both items before entry therefore neither guarantees
nor is required for warrior success.

\begin{table}[p]
\centering\scriptsize
\setlength{\tabcolsep}{3pt}
\caption{Success rates (\%) for the 18 official Intermediate achievements.
Each cell is
an equal-weight mean over three training seeds. The upper panel gives six
current matched cohorts; PPO denotes fresh native GTrXL runs. The lower
panel gives the three D4 intervention arms and historical guard-on PPO
cohorts, which are separate training cohorts from the upper panel.
D4-AH is the fresh joint intervention. M4 marks the four deep-floor
achievements. The source key \texttt{defeat\_orc\_solider} retains its
original spelling.}
\label{tab:tier2-skills}
\begin{tabular}{P{3.35cm}crrrrrr}
\toprule
\multicolumn{8}{l}{\emph{Current matched cohorts}} \\
Achievement & M4 & \shortstack{L40S AHC\\7600} & \shortstack{L40S AHC\\15300} & \shortstack{A100 AHC\\7600} & \shortstack{A100 AHC\\15300} & \shortstack{Fresh PPO\\7600} & \shortstack{Fresh PPO\\15300} \\
\midrule
\texttt{collect\_sapphire} &  & 12.22 & 21.07 & 13.46 & 21.62 & 11.10 & 18.23 \\
\texttt{collect\_ruby} &  & 12.04 & 21.28 & 13.53 & 21.65 & 11.31 & 17.96 \\
\texttt{make\_diamond\_pickaxe} &  & 1.76 & 3.18 & 2.01 & 3.26 & 1.51 & 2.56 \\
\texttt{make\_diamond\_sword} &  & 1.92 & 4.06 & 2.17 & 4.23 & 1.65 & 2.95 \\
\texttt{make\_iron\_armour} &  & 15.71 & 30.98 & 18.25 & 30.33 & 4.33 & 8.02 \\
\texttt{make\_diamond\_armour} &  & 0.0065 & 0.02 & 0.0065 & 0.02 & 0.00 & 0.0098 \\
\texttt{enter\_gnomish\_mines} & $\checkmark$ & 7.57 & 26.38 & 8.05 & 23.88 & 0.48 & 5.43 \\
\texttt{enter\_dungeon} &  & 83.70 & 95.65 & 88.05 & 96.17 & 85.31 & 95.82 \\
\texttt{defeat\_gnome\_warrior} & $\checkmark$ & 1.51 & 20.25 & 1.30 & 8.80 & 0.00 & 0.00 \\
\texttt{defeat\_gnome\_archer} & $\checkmark$ & 1.02 & 13.80 & 0.59 & 6.84 & 0.00 & 0.0033 \\
\texttt{defeat\_orc\_solider} &  & 67.37 & 87.46 & 72.67 & 88.48 & 60.13 & 82.75 \\
\texttt{defeat\_orc\_mage} &  & 52.32 & 76.64 & 55.58 & 77.47 & 42.85 & 69.61 \\
\texttt{eat\_bat} & $\checkmark$ & 1.52 & 18.99 & 0.95 & 8.40 & 0.00 & 0.0033 \\
\texttt{eat\_snail} &  & 74.96 & 90.62 & 79.66 & 91.44 & 74.55 & 89.29 \\
\texttt{find\_bow} &  & 81.34 & 94.51 & 86.07 & 95.10 & 82.39 & 94.50 \\
\texttt{fire\_bow} &  & 77.59 & 93.66 & 83.47 & 94.30 & 75.94 & 92.05 \\
\texttt{open\_chest} &  & 81.34 & 94.51 & 86.07 & 95.10 & 82.39 & 94.50 \\
\texttt{drink\_potion} &  & 53.14 & 74.22 & 58.15 & 74.82 & 53.27 & 73.11 \\
\bottomrule
\end{tabular}

\medskip
\begin{tabular}{P{3.35cm}crrrrrr}
\toprule
\multicolumn{8}{l}{\emph{Component interventions and historical guard-on cohorts}} \\
Achievement & M4 & D4-A & D4-H & D4-AH & PPO end & PPO 7000 & PPO 8000 \\
\midrule
\texttt{collect\_sapphire} &  & 20.92 & 24.10 & 20.73 & 17.27 & 10.51 & 11.95 \\
\texttt{collect\_ruby} &  & 21.22 & 24.06 & 21.08 & 17.15 & 10.47 & 12.21 \\
\texttt{make\_diamond\_pickaxe} &  & 3.16 & 3.63 & 3.06 & 2.48 & 1.46 & 1.72 \\
\texttt{make\_diamond\_sword} &  & 3.92 & 4.84 & 4.03 & 2.57 & 1.57 & 1.73 \\
\texttt{make\_iron\_armour} &  & 26.74 & 22.60 & 30.26 & 6.25 & 4.36 & 6.43 \\
\texttt{make\_diamond\_armour} &  & 0.02 & 0.0098 & 0.0033 & 0.0065 & 0.00 & 0.00 \\
\texttt{enter\_gnomish\_mines} & $\checkmark$ & 22.71 & 30.79 & 18.73 & 4.73 & 0.47 & 0.74 \\
\texttt{enter\_dungeon} &  & 95.95 & 95.97 & 95.93 & 95.22 & 83.17 & 91.16 \\
\texttt{defeat\_gnome\_warrior} & $\checkmark$ & 16.05 & 15.99 & 9.66 & 0.00 & 0.0033 & 0.00 \\
\texttt{defeat\_gnome\_archer} & $\checkmark$ & 10.03 & 11.82 & 7.80 & 0.0065 & 0.00 & 0.00 \\
\texttt{defeat\_orc\_solider} &  & 87.82 & 90.10 & 85.99 & 81.63 & 60.54 & 67.41 \\
\texttt{defeat\_orc\_mage} &  & 76.18 & 82.22 & 74.04 & 68.20 & 42.06 & 47.63 \\
\texttt{eat\_bat} & $\checkmark$ & 14.46 & 15.94 & 9.84 & 0.02 & 0.0033 & 0.0098 \\
\texttt{eat\_snail} &  & 90.86 & 92.03 & 90.81 & 88.80 & 73.63 & 81.00 \\
\texttt{find\_bow} &  & 94.76 & 95.13 & 94.90 & 93.81 & 80.28 & 88.11 \\
\texttt{fire\_bow} &  & 93.77 & 94.47 & 94.04 & 91.09 & 73.74 & 81.96 \\
\texttt{open\_chest} &  & 94.76 & 95.13 & 94.90 & 93.81 & 80.28 & 88.11 \\
\texttt{drink\_potion} &  & 74.71 & 79.77 & 74.59 & 71.29 & 50.97 & 58.62 \\
\bottomrule
\end{tabular}
\end{table}

\subsection{Supplementary checkpoints and training records}
\label{apx:supplement}
The supplemental archive contains Orbax checkpoint data for each primary
training seed at both evaluated budgets (Table~\ref{tab:checkpoint-inventory}),
with file hashes matching the archive contents.

\begin{table}[H]
\centering\small
\caption{Archived AHC checkpoints. Entries count files covered by
the supplied checksum manifest, not independent models or evaluations.
There are three training executions and two checkpoints per execution.}
\label{tab:checkpoint-inventory}
\begin{tabular}{P{3.0cm}P{5.4cm}P{5.4cm}}
\toprule
Training seed & Update 7600: 996,147,200 transitions & Update 15300: 2,005,401,600 transitions \\
\midrule
0 & Present; 12 files & Present; 12 files \\
1 & Present; 12 files & Present; 11 files \\
2 & Present; 11 files & Present; 11 files \\
\bottomrule
\end{tabular}
\end{table}

For each primary run, the training export contains 15,300 consecutive update
records (indices 0--15299) and 151 session-evaluation records. The six logged
update metrics contain no NaN or infinite values. Session evaluations use
changing RNG and episode counts; the final records contain 672, 670, and 692
episodes for seeds 0, 1, and 2, respectively. These training diagnostics
are not used as held-out results.

\subsection{Native-PPO stopping, provenance, and accounting}
\label{apx:ppo-stopped}
The three runs are fresh starts of the \texttt{native\_ppo} study arm,
with training seeds 0--2. Every task record contains only
\texttt{original\_craftax} with probability 1 and a $1\times67$ indicator.
This branch bypasses curriculum task construction and makes no LLM query or
embedding call.

\paragraph{Common configuration and controlled differences.}
Table~\ref{tab:shared-training} records the common student/PPO settings from
the three primary-AHC training configurations, six component-ablation
training configurations, three joint-arm resolved launch configurations,
and three fresh native-PPO training configurations. The additional AHC
cohort uses the same configuration.

\begin{table}[H]
\centering\small
\caption{Shared recorded training settings. These are training parameters;
the frozen evaluation uses its separately specified protocol.}
\label{tab:shared-training}
\begin{tabular}{P{7.0cm}P{6.4cm}}
\toprule
Parameter & Value \\
\midrule
GTrXL layers / attention heads & 2 / 8 \\
Embedding / QKV / hidden width; activation & 256 / 256 / 256; ReLU \\
Transformer gating; gate bias & Enabled; 2.0 \\
Memory / gradient window (steps) & 128 / 64 \\
Task conditioning; action set & 67-dimensional achievement indicator; 43 actions \\
Parallel environments / rollout steps & 1024 / 128 \\
PPO epochs / minibatches & 4 / 8 \\
Discount $\gamma$ / GAE $\lambda$ & 0.999 / 0.8 \\
PPO clipping / entropy / value-loss coefficients & 0.2 / 0.002 / 0.5 \\
Optimizer / Adam epsilon / gradient norm cap & Adam / $10^{-5}$ / 1.0 \\
Learning-rate start / floor / linear horizon & $2\!\times\!10^{-4}$ / $2\!\times\!10^{-6}$ / 15,400 updates \\
GAE value-target clipping & $[-50,300]$ \\
Training episode cap; optimistic reset ratio & 4096; 16 (enabled) \\
Session length / scoring window & 100 / final 40 PPO updates \\
Initialization / full-budget accepted updates & 200 / 15,300 \\
\bottomrule
\end{tabular}
\end{table}

\begin{table}[H]
\centering\small
\caption{Controlled training differences. A/H abbreviate difficulty
annealing and harvested-progress initialization. Removing H retains
per-unit task operators and variant prefills. Historical guard-stopped PPO
is a separate cohort.}
\label{tab:training-differences}
\begin{tabular}{P{2.4cm}P{2.6cm}P{2.4cm}P{3.4cm}P{3.2cm}}
\toprule
Arm & Difficulty A & Progress H & Task / completion reward & Entropy stop \\
\midrule
Full AHC & Adaptive $\sigma$ & Enabled & Curriculum + native anchor; +20 on curriculum success & 0.15 \\
D4-A & Fixed native dynamics & Enabled & Same task interface & 0.15 \\
D4-H & Adaptive $\sigma$ & Neutral floor-zero signature & Same task interface & 0.15 \\
D4-AH & Fixed native dynamics & Neutral floor-zero signature & Same task interface & 0.15 \\
Fresh native PPO & Native dynamics & Native resets & Original task only; native reward & Disabled \\
\bottomrule
\end{tabular}
\end{table}

All arms retain the value-loss ceiling of 1000 and the numerical
guardrails. Fresh native PPO disables only the entropy stop and keeps the
entropy regularization coefficient of 0.002. The comparison holds the
student and PPO parameters fixed while changing the training procedure.

\paragraph{Historical guard-stopped cohort.}
The configurations use 1024 environments, 128 steps per update, four PPO
epochs, and eight minibatches. The learning-rate settings are
$2\times10^{-4}$ initially, a floor of $2\times10^{-6}$, and a schedule
horizon of 15,400 updates, while the planned run length is 15,300 updates.
The session guard uses entropy threshold 0.15 and value-loss threshold 1000.
For seeds 0--2, the rejected sessions begin at updates 14600, 15000,
and 13600. Their minimum logged entropies are 0.147797, 0.138487, and
0.149899, respectively. The rejected session is reverted and training
stops; Table~\ref{tab:ppo-stopped} uses the last accepted checkpoint.

Evaluation uses Craftax 1.4.5, Python 3.11.15, JAX/JAXlib 0.6.2, Flax
0.10.7, and Orbax 0.11.18, with evaluation seed 0, 1024 first episodes, an
8192-step cap, unmasked actions, and the 67-dimensional all-ones native-task
indicator.

\begin{table}[H]
\centering\small
\caption{Single-rollout measurements for the three guard-stopped PPO
endpoints (first episode on each of 1024 worlds). Performance is return as a
percentage of 226. Training time is elapsed time on one RTX A6000,
including in-run evaluation and rejected sessions.}
\label{tab:ppo-stopped-accounting}
\begin{tabular}{rrrrr}
\toprule
Seed & Return & Performance (\%) & Mean episode length & Training RTX A6000 GPU$\cdot$h \\
\midrule
0 & 41.843 & 18.514706 & 1337.300 & 22.4089 \\
1 & 40.668 & 17.994696 & 1351.456 & 22.9050 \\
2 & 42.068 & 18.614149 & 1159.646 & 20.8015 \\
\bottomrule
\end{tabular}
\end{table}

Under this single-rollout protocol, return is $41.526\pm0.752$,
performance is $18.375\pm0.333$\%, and episode length is
$1282.801\pm106.889$ (sample SD). Mine entry is $5.273\pm3.565$\%
(8.887\%, 1.758\%, and 5.176\% for seeds 0--2), warrior and archer success
are both zero for every seed, and bat success is $0.033\pm0.056$\%.
Training totals 66.1155 RTX A6000 GPU$\cdot$h. These runs are separate from
the fresh full-budget runs described below.

\paragraph{Full-budget runs.}
The fresh runs retain training seeds 0--2, 1024 environments, 128 steps per update, four PPO epochs, eight minibatches, and the original student, optimizer, environment, and learning-rate schedule. The session entropy stopping threshold is set to $-\infty$; the PPO entropy regularization coefficient remains 0.002. The numerical checks reject non-finite \texttt{value\_loss} and a mean value loss above 1000.
All three runs are fresh starts of \texttt{native\_ppo} with \texttt{original\_craftax} as the only task and the all-ones achievement indicator. The accepted update records cover 0--15,299, reaching 2,005,401,600 PPO rollout transitions. All three train on RTX A6000, with occupied-process durations of 23.2864, 23.2814, and 23.3934 GPU$\cdot$h (69.9612 in total). We evaluate the checkpoints at updates 7600 and 15300. Seed labels are reused from the historical arm but identify new executions.

\subsection{D4 ablation records and intervention scope}
\label{apx:d4}
The D4-A and D4-H arms each have three training runs of 151 curriculum
sessions, 15,300 PPO updates, and 2,005,401,600 rollout transitions, trained
on RTX A6000 with Python 3.11.15, JAX/JAXlib 0.6.2, and Craftax 1.4.5.
Table~\ref{tab:d4-inputs} gives their single-rollout results; repeated-world
results are in Appendix~\ref{apx:unified}.

\begin{table}[H]
\centering\small
\caption{D4 final-checkpoint results by training seed. These are individual
run means over 1024 first episodes. Performance is return as a percentage
of the 226-point maximum.}
\label{tab:d4-inputs}
\begin{tabular}{llrrr}
\toprule
Intervention & Seed & Return & Performance (\%) & Episode length \\
\midrule
Fixed difficulty & 0 & 47.812920 & 21.156158 & 1462.630 \\
Fixed difficulty & 1 & 45.908348 & 20.313429 & 1553.788 \\
Fixed difficulty & 2 & 45.132290 & 19.970039 & 1457.445 \\
No progress initialization & 0 & 46.757309 & 20.689075 & 2115.051 \\
No progress initialization & 1 & 44.585232 & 19.727980 & 1216.120 \\
No progress initialization & 2 & 48.070557 & 21.270159 & 1616.744 \\
\bottomrule
\end{tabular}
\end{table}

Across seeds, D4-A performance is $20.480\pm0.610$\% and episode length is
$1491.288\pm54.189$; D4-H gives $20.562\pm0.779$\% and
$1649.305\pm450.349$, respectively (sample SD). D4-A retains the four-skill boundary in all three runs; D4-H retains it in two.

The intervention is selected through \texttt{AHC\_STUDY\_ARM}.
For D4-A, \texttt{anneal/curator.py} guards changes to $\sigma$ while retaining
the other controller branches. All three training summaries record zero
sigma moves. For D4-H, \texttt{study.initialization\_signature} substitutes
\texttt{(0, (), (), (), ())} at state realization in
\texttt{anneal/frontier.py}; bank admission remains intact. Variant-defined
prefills, resource and lighting operations, and independent
variants are retained. The intervention therefore removes harvested
signature inheritance while keeping the other initial-state operators.

\paragraph{Joint intervention.}
The joint arm has three fresh training runs (seeds 0--2) with both the D4-A
and D4-H interventions active. Each run completes 151 sessions and 15,300
updates (2,005,401,600 rollout transitions) with zero sigma moves. The
student and optimizer settings, bank admission, variant-defined prefills,
selector, anchor, and other state operations are unchanged.

\paragraph{Joint execution source.}
The joint arm extends \texttt{study.py} so that one arm activates both
intervention predicates, and registers it in \texttt{launch.py}. In
\texttt{anneal/session.py}, the floor-zero world bucket that D4-H already
ensures is extended to any arm that clears signatures, so the joint arm uses
the same floor-zero provision rather than a new pool intervention. Both arms
use the same base world-pool asset, and the pool generator, sampler, and
state-assembly files are identical.

Each joint checkpoint is evaluated with the same repeated-world scripts
(10,240 episodes per seed); per-seed results and all 18 Intermediate skill
means are in Tables~\ref{tab:unified-seeds} and~\ref{tab:tier2-skills}.

A separate fixed-difficulty seed-1 run trained on A100 returns 46.290 under the single-rollout protocol, with mine entry, warrior, archer, and bat rates of 24.61\%, 18.26\%, 12.40\%, and 17.09\%, retaining the four-skill boundary; its repeated-world return is 46.20.

\subsection{Collaborator DiCode reproduction summary}
\label{apx:dicode-repro}
The supplied summary reports three training seeds at checkpoint 15300 and
states that the AHC comparison uses the same 1024 worlds. Its return mean
and sample SD are $48.77\pm2.10$; a separate comparison in that summary uses
SE ($48.77\pm1.21$ for DiCode and $46.75\pm0.33$ for AHC).
Table~\ref{tab:dicode-repro} uses sample SD at the reported precision.

\begin{table}[H]
\centering\small
\caption{DiCode reproduction capability profile at checkpoint 15300,
transcribed from the collaborator summary. Rates are percentages; the last
row is the supplied across-seed mean.}
\label{tab:dicode-skills}
\begin{tabular}{lrrrr}
\toprule
Training seed & Mine entry & Warrior & Archer & Bat \\
\midrule
0 & 22.4 & 0.1 & 0.1 & 0.0 \\
1 & 41.9 & 36.8 & 29.9 & 36.4 \\
2 & 34.7 & 26.8 & 15.3 & 25.2 \\
\midrule
Reported mean & 33.0 & 21.2 & 15.1 & 20.5 \\
\bottomrule
\end{tabular}
\end{table}

Two of the three DiCode runs exceed the four-skill boundary. Endpoint JSONs
for seeds 1 and 2 give returns 50.929611 and 48.636997, agreeing with the
rounded values in Table~\ref{tab:dicode-repro}. These runs use
\texttt{qwen3-235b-a22b-thinking-2507} for generation and
\texttt{Qwen/Qwen3-Embedding-0.6B} for embeddings, upstream revision
\texttt{35d4ae4} with modifications, Craftax 1.4.5, and JAX 0.6.2. This
teacher differs from both the 14B diagnostic teacher and the published 80B
comparison.

\subsection{External version and reporting boundaries}
\label{apx:version}
The experiments in this study use Craftax 1.4.5. On identical checkpoints,
Craftax 1.6.1 gave readings 0.23--0.29 points higher in two runs; we do not
apply this offset to external values, since DiCode's dependency declaration
does not pin a Craftax version. The DiCode numbers use the project-page
eight-seed set rather than the earlier arXiv five-seed set.

The PPO-GTrXL 4B conversion uses $0.206\times226=46.556$, rounded to 46.56 \citep{hamon2024transformerxl}. External GTrXL values are taken as reported.

\subsection{Diagnostic interventions and graph-exemption timing}
\label{apx:graph-timing}

\begin{table}[H]
\centering\small
\caption{Diagnostic interventions relative to longStack. All three
training seeds in each arm use the same arm definition. This historical
diagnostic comparison is separate from the AHC component ablations.}
\label{tab:diagnostic-interventions}
\begin{tabular}{P{2.0cm}P{5.1cm}P{6.8cm}}
\toprule
Arm & Graph change & Scheduler, gate, and prompt change \\
\midrule
longStack & Original 67-node, 94-edge graph & Prerequisite scheduler; R3 floor clause enabled \\
repairF & Replace mine-entry prerequisite with dungeon entry for warrior, archer, and bat only & Scheduler and R3 use repaired graph; mastered-prerequisite exemption retained; API-reference content differs \\
shufGraph & Fixed double-edge rewiring, seed 20260804; 79 of 94 original edges replaced & Scheduler and R3 use shuffled graph; target identities and scaffold checks both change \\
Surgical & Original graph retained & Only floor-granted achievements removed from R3 scaffold checks; premark/inventory channels and R1/R2 retained \\
\bottomrule
\end{tabular}
\end{table}

\paragraph{Graph construction.}
RepairF replaces each of the three edges
$\texttt{enter\_gnomish\_mines}\to
\{\texttt{defeat\_gnome\_warrior},\texttt{defeat\_gnome\_archer},
\texttt{eat\_bat}\}$ with the corresponding
$\texttt{enter\_dungeon}$ edge; all other relations are unchanged.
The original, repaired, and shuffled graph fingerprints are
\texttt{cfbb1c9a4558}, \texttt{202b5f536b37}, and
\texttt{1354f4e59b14}. Shuffling starts from the sorted directed edge
list and Python's \texttt{random.Random(20260804)}. Draw two edge indices
with replacement; for $(a,b),(c,d)$, require four distinct endpoints and
no existing $(a,d)$ or $(c,b)$ edge. Tentatively swap their destinations
and reject the swap if the graph becomes cyclic. Stop after $20|E|$
accepted swaps or $400|E|$ attempts. The retained launch logs report
1880 accepted swaps in 2566 attempts, 67 nodes, 94 edges, unchanged
in/out-degree sequences, and 79 original edges absent from the result.
This is one fixed permutation across training seeds, not three independently
sampled graphs.

\paragraph{The two graph consumers.}
For scheduler input rates $s_a$ (the retained max-ever mastery readings),
a target satisfies $s_a<0.60$ and $s_p\geq0.30$ for \emph{every} direct
prerequisite $p$. Eligible targets are sorted by increasing success rate,
then decreasing fixed achievement tier, then name; up to six are issued. If none is
eligible, the scheduler falls back to its tier rule. Rewiring changes the
relations tested, not the numerical 0.30 threshold.
The scaffold gate's R3 instead considers relevant targets with rate below
0.70. It flags any immediate prerequisite granted by premarking,
inventory, or starting floor; when the mastered-prerequisite exemption is
enabled, a prerequisite already at or above 0.70 is exempt. The retained
configurations enable this exemption for longStack, repairF, shufGraph,
and Surgical. Surgical removes only the starting-floor channel. RepairF
changes the graph while retaining the exemption; its API-reference setting
also differs from longStack. The timing record below describes the observed
gate events.

\begin{figure}[H]
\centering
\includegraphics[width=\linewidth]{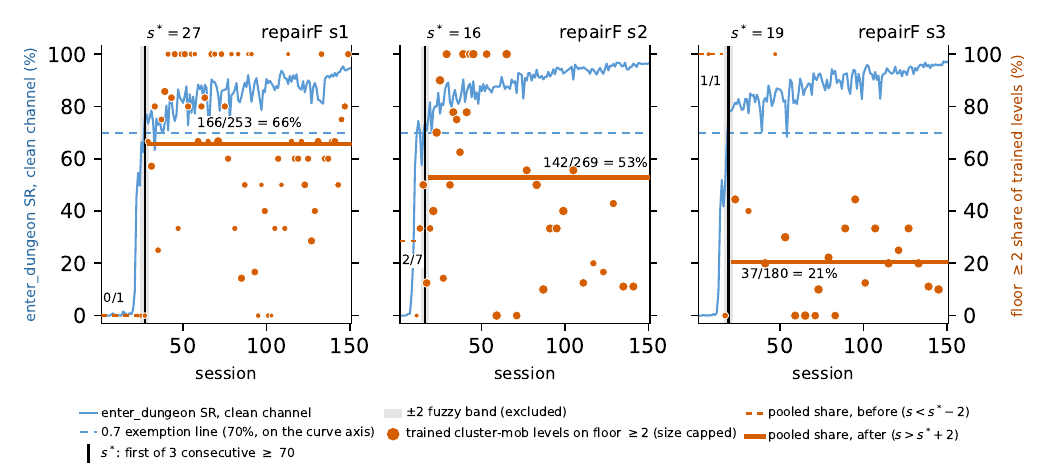}
\caption{Mastered-prerequisite exemption timing in the three repairF runs.
The blue curve is the clean-channel \texttt{enter\_dungeon} success rate.
Black lines mark $s^*$, the first of three consecutive readings at or above
70\% (the 0.7 criterion): sessions 27, 16, and 19, respectively. Orange dots
give the proportion of cluster-mob levels whose generated code starts on
floor 2 or deeper, among levels with a nonempty training-performance history.
Levels target gnome warriors, gnome archers, or bats and are grouped by their
creation session, rather than by the session when they were trained. Dot size
increases with the number of counted levels and is capped. Orange segments
pool level counts before ($s < s^*-2$) and after ($s > s^*+2$), excluding the
inclusive $s^*\pm2$ band. Before/after numerator--denominator pairs are
0/1 and 166/253, 2/7 and 142/269, and 1/1 and 37/180.}
\label{fig:door}
\end{figure}

\subsection{Native PPO with entropy-guard stopping}
\label{sec:ppo-stopped}
The native-PPO branch trains the same GTrXL student from scratch using only
the original Craftax task and a constant 67-dimensional all-achievement
indicator. It invokes neither the curriculum controller nor LLM generation
or embedding. The supplied run configurations target 15,300 updates, but all
three original runs stop when a session's minimum logged entropy falls
below the configured threshold of 0.15. That session is reverted and the
study driver terminates; the retained checkpoint is the last accepted
training state, not the output of the rejected session.

Table~\ref{tab:ppo-stopped} reports those retained endpoints at their actual
budgets with the repeated-world protocol used for AHC: ten rollouts on each
of the 1024 worlds, an 8192-step cap, and the native unmasked action space.
Their original single-rollout evaluations are retained in
Appendix~\ref{apx:ppo-stopped}.

\begin{table}[H]
\centering\small
\setlength{\tabcolsep}{4pt}
\caption{Original native-PPO runs stopped by the entropy guard, evaluated
with ten rollouts on each of 1024 worlds. Updates and PPO rollout
transitions describe the last valid checkpoint; each transition count
equals updates $\times1024\times128$. Skill rates are percentages of 10,240
episodes. Every run stops short of the 2,005,401,600-transition target.}
\label{tab:ppo-stopped}
\begin{tabular}{rrrrrrrr}
\toprule
Seed & Updates & Valid transitions & Return & Mine entry & Warrior & Archer & Bat \\
\midrule
0 & 14600 & 1,913,651,200 & 42.086 & 7.266 & 0 & 0 & 0.029 \\
1 & 15000 & 1,966,080,000 & 40.249 & 2.275 & 0 & 0 & 0 \\
2 & 13600 & 1,782,579,200 & 41.867 & 4.648 & 0 & 0.020 & 0.029 \\
\bottomrule
\end{tabular}
\end{table}

The stopped endpoints average $41.401\pm1.003$ (sample SD), 5.30 points
below AHC's repeated-world mean of 46.700 (Welch 95\% CI $[3.24, 7.36]$);
their single-rollout mean was $41.526\pm0.752$. None retains the four-skill boundary: mine entry ranges from 2.28\% to 7.27\%, no episode
records a warrior success, and archer and bat successes each occur in at
most three of 10,240 episodes per run.

Because these runs stop short of the full budget, the main comparison uses a separate fresh native-PPO cohort that completes the full rollout budget with only the session entropy stop disabled and reaches $41.91\pm1.01$ (\secref{sec:pending}). Appendix~\ref{apx:ppo-stopped} gives the training and evaluation records of the historical endpoints.

\subsection{Three-seed DiCode reproduction}
\label{sec:dicode-repro}
A collaborator's DiCode reproduction reports final-checkpoint returns of
46.74, 50.93, and 48.64 for training seeds 0--2. Table~\ref{tab:dicode-repro}
compares these with the existing AHC single-rollout results at checkpoint
15300. The supplied summary states that both use the same 1024 worlds.

\begin{table}[H]
\centering\small
\caption{Three-seed endpoint returns on the reported common 1024-world
evaluation. AHC values come from the original single-rollout JSONs; DiCode
values and its sample SD are reported at the supplied summary's precision.}
\label{tab:dicode-repro}
\begin{tabular}{lrrrr}
\toprule
System & seed 0 & seed 1 & seed 2 & Mean $\pm$ sample SD \\
\midrule
AHC (single rollout) & 46.861 & 47.268 & 46.134 & $46.754\pm0.574$ \\
DiCode (collaborator reproduction) & 46.74 & 50.93 & 48.64 & $48.77\pm2.10$ \\
\bottomrule
\end{tabular}
\end{table}

AHC's single-rollout mean is about 2.02 points lower than the supplied
DiCode mean, with a smaller spread across seeds. DiCode's capability profile
also varies across seeds: seed 0 has 22.4\% mine entry but only 0.1\%
warrior and archer success and no bat success
(Appendix~\ref{apx:dicode-repro}).

\subsection{Per-run diagnostic capability rates}
\label{apx:rates}
The following tail-window table gives the underlying skill rates for the
diagnostic figures and recovery counts. It is not pooled with the
final-endpoint AHC comparison. Table~\ref{tab:perseed-final} gives the
checkpoint-15300 readings of the same runs, which is DiCode's reporting
convention; every recovery label is unchanged under that reading.

\begin{table}[H]
\centering\footnotesize
\caption{Per-seed last-10 means (\%) for the four second-floor skills. Supports the base15--longStack separation and the recovery counts; other recorded configurations are included for completeness.}
\label{tab:perseed}
\begin{tabular}{llrrrr}
\toprule
configuration & seed & mines & warrior & archer & eat\_bat \\
\midrule
shufGraph & s1 & 19.06 & 14.97 & 11.00 & 15.05 \\
 & s2 & 26.94 & 22.00 & 13.89 & 21.02 \\
 & s3 & 26.33 & 21.13 & 13.50 & 19.38 \\
\addlinespace[1pt]
repairF & s1 & 19.73 & 14.36 & 5.15 & 5.83 \\
 & s2 & 23.37 & 18.17 & 12.66 & 15.78 \\
 & s3 & 29.30 & 24.06 & 18.73 & 22.67 \\
\addlinespace[1pt]
surgical & s1 & 18.60 & 14.79 & 6.32 & 11.09 \\
 & s2 & 15.48 & 11.74 & 3.95 & 7.84 \\
  & s3 & 19.65 & 15.62 & 10.90 & 14.51 \\
\addlinespace[1pt]
base2e9 & s1 & 17.14 & 13.53 & 6.86 & 12.72 \\
 & s2 & 20.68 & 16.45 & 10.92 & 15.83 \\
 & s3 & 24.33 & 20.12 & 15.03 & 21.27 \\
\addlinespace[1pt]
longStack & s1 & 10.26 & 4.54 & 0.28 & 0.88 \\
 & s2 & 8.00 & 0.12 & 0.00 & 0.62 \\
 & s3 & 10.66 & 0.00 & 0.00 & 0.00 \\
\addlinespace[1pt]
base15 & s1 & 23.19 & 19.01 & 12.45 & 18.25 \\
 & s2 & 15.31 & 11.82 & 8.08 & 11.71 \\
 & s3 & 18.81 & 14.52 & 6.34 & 12.27 \\
\addlinespace[1pt]
gateOff & s1 & 15.85 & 12.46 & 3.80 & 11.41 \\
 & s2 & 14.19 & 0.06 & 0.00 & 0.02 \\
 & s3 & 6.39 & 0.02 & 0.01 & 0.00 \\
\addlinespace[1pt]
breadth08 & n=1 & 7.53 & 0.01 & 0.00 & 0.05 \\
\addlinespace[1pt]
cap300 & n=1 & 18.55 & 14.73 & 10.47 & 15.28 \\
\bottomrule
\end{tabular}
\end{table}

\begin{table}[H]
\centering\footnotesize
\caption{Per-run readings at checkpoint 15300 from the same evaluation
summaries as Table~\ref{tab:perseed}: return in points (with the last-ten mean
beside it) and the four second-floor skill rates in \%. Recovery labels are
unchanged from the tail window.}
\label{tab:perseed-final}
\begin{tabular}{llrrrrrr}
\toprule
configuration & seed & return & last-ten & mines & warrior & archer & eat\_bat \\
\midrule
shufGraph & s1 & 45.480 & 45.661 & 20.21 & 16.02 & 12.11 & 15.62 \\
 & s2 & 47.652 & 46.520 & 29.00 & 25.10 & 16.60 & 23.83 \\
 & s3 & 47.848 & 47.041 & 29.98 & 24.41 & 15.23 & 22.75 \\
\addlinespace[1pt]
repairF & s1 & 44.908 & 44.368 & 22.85 & 16.11 & 4.88 & 7.23 \\
 & s2 & 45.980 & 46.270 & 22.36 & 16.21 & 11.82 & 15.33 \\
 & s3 & 48.484 & 47.328 & 33.50 & 27.44 & 21.39 & 26.37 \\
\addlinespace[1pt]
surgical & s1 & 46.916 & 44.857 & 23.93 & 21.00 & 10.94 & 15.92 \\
 & s2 & 46.047 & 45.258 & 21.09 & 16.70 & 5.47 & 11.13 \\
  & s3 & 46.878 & 46.305 & 21.97 & 17.29 & 13.09 & 16.70 \\
\addlinespace[1pt]
base2e9 & s1 & 43.762 & 43.115 & 17.97 & 15.04 & 9.47 & 14.65 \\
 & s2 & 45.344 & 43.998 & 24.22 & 19.24 & 13.67 & 19.14 \\
 & s3 & 46.365 & 45.650 & 27.93 & 22.75 & 16.02 & 23.83 \\
\addlinespace[1pt]
longStack & s1 & 45.016 & 44.726 & 12.40 & 6.74 & 0.20 & 1.46 \\
 & s2 & 43.545 & 43.157 & 8.01 & 0.00 & 0.00 & 0.98 \\
 & s3 & 45.206 & 44.159 & 14.94 & 0.00 & 0.00 & 0.00 \\
\addlinespace[1pt]
base15 & s1 & 46.024 & 45.016 & 24.80 & 21.09 & 14.55 & 20.51 \\
 & s2 & 42.549 & 42.476 & 13.96 & 11.52 & 7.71 & 10.64 \\
 & s3 & 44.955 & 44.296 & 21.19 & 17.38 & 7.62 & 14.06 \\
\addlinespace[1pt]
gateOff & s1 & 45.375 & 45.061 & 17.29 & 13.09 & 5.18 & 13.48 \\
 & s2 & 44.883 & 44.419 & 16.02 & 0.00 & 0.00 & 0.00 \\
 & s3 & 41.289 & 40.730 & 7.81 & 0.10 & 0.00 & 0.00 \\
\addlinespace[1pt]
breadth08 & n=1 & 43.164 & 43.453 & 7.81 & 0.00 & 0.00 & 0.00 \\
\addlinespace[1pt]
cap300 & n=1 & 42.567 & 41.756 & 20.90 & 16.80 & 12.50 & 18.07 \\
\bottomrule
\end{tabular}
\end{table}

\subsection{Historical and external return context}
\label{apx:return-context}
These records place the internal comparison in
Table~\ref{tab:internal-budget} in context.

\begin{table}[H]
\centering\scriptsize
\setlength{\tabcolsep}{4pt}
\caption{Historical single-rollout final-endpoint comparison, recomputed from existing
checkpoint-15300 JSON entries. Each cell is mean $\pm$ sample SD over three
training runs. All rows use Craftax 1.4.5, evaluation seed 0, and 1024
environments with one rollout per world and an 8192-step cap. The AHC row is
the original single-rollout result, not Table~\ref{tab:mechanical-endpoint}'s
repeated-world estimate. Skill columns are percentages.}
\label{tab:local-endpoint}
\begin{tabular}{lccccc}
\toprule
System & Return & Mine entry & Warrior & Archer & Bat \\
\midrule
AHC & $46.754 \pm 0.574$ & $26.758 \pm 2.633$ & $20.768 \pm 2.274$ & $14.225 \pm 0.853$ & $19.206 \pm 0.451$ \\
base15 & $44.509 \pm 1.780$ & $19.987 \pm 5.519$ & $16.667 \pm 4.825$ & $9.961 \pm 3.975$ & $15.072 \pm 5.008$ \\
longStack & $44.589 \pm 0.909$ & $11.784 \pm 3.508$ & $2.246 \pm 3.890$ & $0.065 \pm 0.113$ & $0.814 \pm 0.746$ \\
\bottomrule
\end{tabular}
\end{table}

Under the original single-rollout protocol at the common endpoint, AHC's
mean return is 2.245 points above base15 and 2.166 above longStack. Its mean
is also higher on each of the four skills. The altered-graph and floor-clause
configurations of \secref{sec:diagnosis} are within 0.3 return points of AHC
under this same single-rollout protocol (Table~\ref{tab:unmasked-frontier}).
Per-run values are in \apxref{apx:measure}.

\paragraph{Historical and external context.}
Guard-stopped native-PPO runs stop short of the target budget
(Appendix~\ref{sec:ppo-stopped}). A collaborator's DiCode reproduction
reports $48.77\pm2.10$, about 2.02 points above AHC's single-rollout mean
(Appendix~\ref{sec:dicode-repro}).

\begin{table}[H]
\centering\small
\caption{External and historical numerical context, separate from the internal
two-budget comparison in Table~\ref{tab:internal-budget}. External results use different
software versions, evaluation protocols, and time caps; DiCode uncertainties
are standard errors \citep{mitsides2026dicode}. The 14B diagnostic panel
uses one rollout on each of 1024 worlds: its checkpoint-15300 mean $\pm$ SD
is followed by the last-ten mean in parentheses. The other two diagnostic
endpoint means are
base2e9 45.16 and gateOff 43.85 (Appendix~\ref{apx:rates}).}
\label{tab:unmasked-frontier}
\setlength{\tabcolsep}{4pt}
\begin{tabular}{P{6.1cm}cccc}
\toprule
Model & Budget (B) & LLM & Mask & Return \\
\midrule
\multicolumn{5}{l}{\emph{Near 1B transitions}} \\
PPO-GTrXL (maintainer board) & 1.000 & No & None & 41.36 \\
\addlinespace
\multicolumn{5}{l}{\emph{Near 2B transitions}} \\
PPO-GTrXL (DiCode report) & 2.000 & No & None & 41.59 \\
DiCode 30B (reported) & 2.000 & Yes & Not audited & $44.38\pm0.66$ \\
DiCode 80B (reported) & 2.000 & Yes & Not audited & $45.88\pm0.40$ \\
DiCode 235B (reported) & 2.000 & Yes & Not audited & $48.55\pm0.49$ \\
\addlinespace
\multicolumn{5}{l}{\emph{14B diagnostic configurations: single-rollout endpoint (tail mean)}} \\
base15 & $\approx2$ & Yes & None & $44.51\pm1.78$ (43.93) \\
longStack & $\approx2$ & Yes & None & $44.59\pm0.91$ (44.01) \\
repairF & $\approx2$ & Yes & None & $46.46\pm1.84$ (45.99) \\
shufGraph & $\approx2$ & Yes & None & $46.99\pm1.31$ (46.41) \\
Surgical & $\approx2$ & Yes & None & $46.61\pm0.49$ (45.47) \\
\addlinespace
\multicolumn{5}{l}{\emph{4B transitions}} \\
PPO-GTrXL (repository) \citep{hamon2024transformerxl} & 4.000 & No & None & 46.56 \\
\bottomrule
\end{tabular}
\end{table}

\paragraph{Teacher-size context.}
The published DiCode ladder reports $44.38\pm0.66$, $45.88\pm0.40$,
and $48.55\pm0.49$ for 30B, 80B, and 235B teachers, respectively
(standard errors). The repeated-world L40S AHC mean is numerically 0.82
points above the reported 80B result and 1.85 below the 235B result.
Under the historical single-rollout protocol, the five 14B diagnostic
means span 44.51--46.99; repairF, shufGraph, and Surgical are numerically
above the reported 30B and 80B means and within 0.3 points of AHC's
single-rollout 46.754. AHC thus reaches a similar return range without LLM
inference.

\paragraph{Return contrasts.}
Table~\ref{tab:return-contrasts} lists the Welch intervals referred to in
the main text, together with the ablation and historical stopped-run
contrasts.

\begin{table}[H]
\centering\small
\caption{Return differences and Welch 95\% intervals from three training-seed means in each cohort, using the ten-repeat evaluator. The final three rows compare against historical guard-stopped endpoints at their own budgets.}
\label{tab:return-contrasts}
\begin{tabular}{P{8.9cm}rc}
\toprule
Comparison & Difference & Welch 95\% interval \\
\midrule
L40S AHC $-$ fresh PPO, update 7600 & $2.06$ & $[-2.56, 6.68]$ \\
A100 AHC $-$ fresh PPO, update 7600 & $3.76$ & $[0.18, 7.33]$ \\
L40S AHC $-$ fresh PPO, update 15300 & $4.80$ & $[2.71, 6.88]$ \\
A100 AHC $-$ fresh PPO, update 15300 & $4.17$ & $[1.58, 6.76]$ \\
D4-A $-$ fresh PPO, update 15300 & $4.20$ & $[1.67, 6.72]$ \\
D4-H $-$ fresh PPO, update 15300 & $4.51$ & $[1.35, 7.67]$ \\
D4-AH $-$ fresh PPO, update 15300 & $3.23$ & $[-2.87, 9.33]$ \\
D4-AH $-$ L40S AHC, update 15300 & $-1.56$ & $[-8.15, 5.02]$ \\
A100 AHC $-$ L40S AHC, update 15300 & $-0.62$ & $[-3.21, 1.96]$ \\
Fresh PPO $-$ historical stopped PPO & $0.50$ & $[-1.78, 2.79]$ \\
D4-A $-$ historical stopped PPO & $4.70$ & $[2.19, 7.21]$ \\
D4-H $-$ historical stopped PPO & $5.02$ & $[1.86, 8.18]$ \\
\bottomrule
\end{tabular}
\end{table}

\end{document}